\documentclass[twoside]{article}

\usepackage{amsmath}
\usepackage{amssymb}
\usepackage{subcaption}
\usepackage{placeins}
\usepackage{booktabs}
\usepackage{multirow} 
\usepackage{graphicx}

\usepackage{algorithm}
\usepackage{algorithmic}
\usepackage{float}

\usepackage[preprint]{aistats2026}
\usepackage[round]{natbib}
\usepackage{hyperref}
\begin{document}

%

%

\twocolumn[

\aistatstitle{RealStats: A Rigorous Real-Only Statistical Framework for Fake Image Detection}

\aistatsauthor{ Haim Zisman \And Uri Shaham }

\aistatsaddress{ Bar-Ilan University \And Bar-Ilan University} 
]

\begin{abstract}
As generative models continue to evolve, detecting AI-generated images remains a critical challenge. While effective detection methods exist, they often lack formal interpretability and may rely on implicit assumptions about fake content, potentially limiting their robustness to distributional shifts. In this work, we introduce a rigorous, statistically grounded framework for fake image detection that focuses on producing a probability score interpretable with respect to the real-image population. Our method leverages the strengths of multiple existing detectors by combining strong training-free statistics. We compute $p$-values over a range of test statistics and aggregate them using classical statistical ensembling to assess alignment with the unified real-image distribution. This framework is generic, flexible, and training-free, making it well-suited for robust fake image detection across diverse and evolving settings.

\noindent Code available at: \url{https://github.com/shaham-lab/RealStats}.

\end{abstract}

\footnotetext{Accepted to AISTATS 2026.}

\section{Introduction}

The ability to distinguish real images from AI-generated content has become increasingly critical as generative models grow in both realism and accessibility. Recent advances in generative AI have significantly improved the visual fidelity of synthetic images, making them nearly indistinguishable from real ones to the human eye.

Addressing this growing challenge requires detection methods that are explicitly designed to balance two key properties central to our approach: interpretability and adaptability. Interpretability refers to the ability to assess the reliability of detection outputs by producing scores with well-defined statistical meaning. This is essential to ensure that the results can be understood and trusted in practice.

Adaptability denotes the ability to remain effective under distribution shifts caused by emerging generators. It requires operating without relying on assumptions about fake image distributions, while allowing future refinement and extension as new detection signals or insights become available.

Detection methods for AI-generated images are rapidly advancing. A major direction has focused on supervision, where models or ensembles are trained on labeled fake images to learn patterns that distinguish real from synthetic content~\cite{wang2020cnn, martin2023pixel, epstein2023online}. 
\begin{figure}[!ht]
    \centering
    \includegraphics[width=0.46\textwidth]{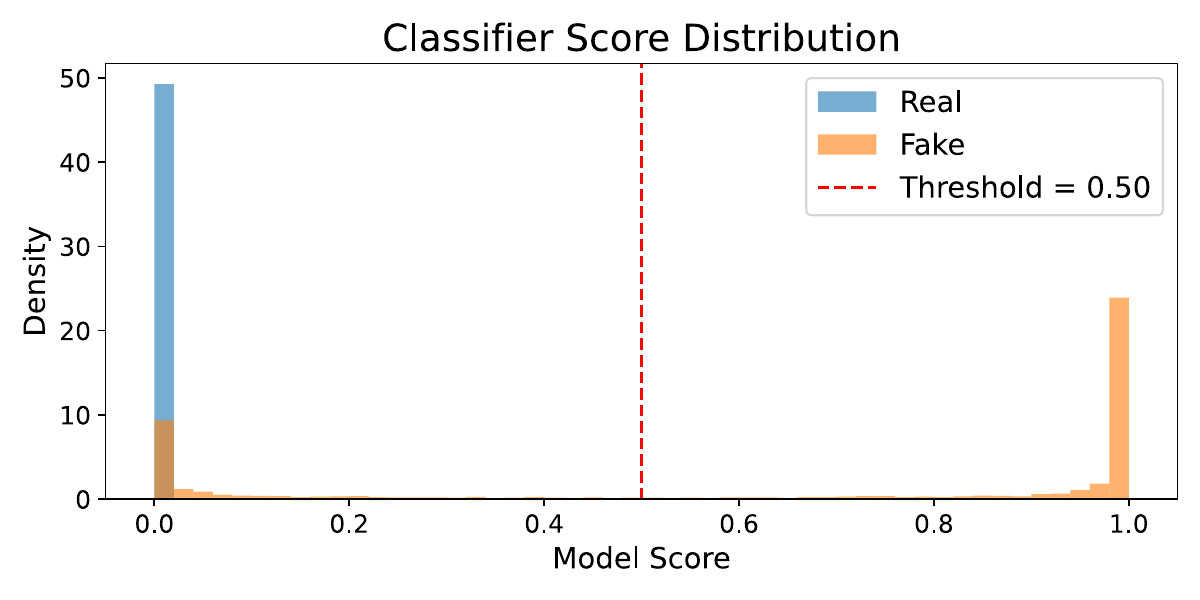}
    \includegraphics[width=0.46\textwidth]{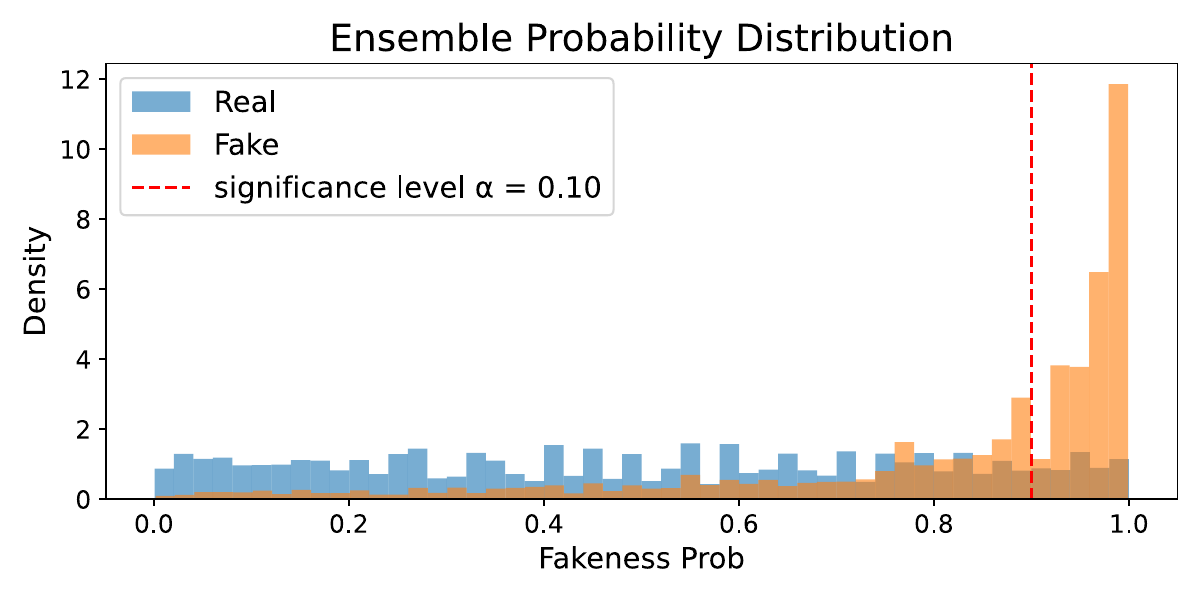}
    \caption{
    Illustration of the score interpretability gap between a supervised classifier~\cite{wang2020cnn} and our statistical method.
    \textbf{Top:} A supervised model outputs scores that can separate real from fake images, but these scores are not inherently interpretable, as they lack clear statistical meaning and are often overconfident.
    \textbf{Bottom:} Our method produces calibrated $p$-values based on real image distributions. These values have a precise interpretation: the probability of observing such a result if the image were real. This enables principled decisions using a standard significance level.
    }

    \label{fig:score_distribution}
\end{figure}
Although effective at capturing training distributions, their reliance on fake data limits adaptability, as performance often degrades under distribution shifts introduced by evolving generators~\cite{gragnaniello2021gan}.

Recent work explores strong pre-trained representations~\cite{ojha2023towards, cozzolino2024raising} and hand-crafted heuristics~\cite{ricker2024aeroblade, he2024rigid, brokman2025manifold} as detection statistics, using real images as a reference. These approaches enhance adaptability by avoiding training on fake data and offer a degree of interpretability by calibrating thresholds based on real image distributions. However, their adaptability remains limited, as the distribution shifts introduced by evolving generators can still affect performance. This may result from assumptions about how fake images differ from real ones, such as fixed magnitude relationships that do not generalize, or from the inability of individual statistics to consistently separate real and synthetic content across different generative models.

We extend recent training-free detection methods by proposing a statistically rigorous framework based solely on real images. To improve adaptability, the method integrates multiple detection statistics to capture a broader range of deviations introduced by evolving generators, without making any assumptions about fake image distributions.

Our method returns a $p$-value under the null hypothesis that an image is drawn from the real distribution, providing a statistically grounded and interpretable measure of reliability. This value is derived by aggregating several intermediate $p$-values, each capturing deviation under a different statistic, into a single coherent score reflecting the overall evidence against the image being real. In summary, \textbf{our contribution} is a real-only detection framework that formulates AI-image detection as a hypothesis test on the real distribution and integrates multiple training-free statistics into a calibrated $p$-value. This yields an interpretable and adaptable structure with formal statistical meaning, allowing the method to incorporate new statistics without relying on fake data.

\section{Related Work}

Efforts to detect AI-generated images have evolved across three main paradigms: fully supervised methods, few-shot and zero-shot detectors leveraging pretrained models, and unsupervised approaches rooted in statistical inference. Each class of methods has contributed different insights into the detection problem.

Supervised approaches are a widely explored direction, training discriminative models on labeled datasets of real and synthetic images. CNN-based methods~\cite{wang2020cnn, baraheem2023ai} have demonstrated strong performance, and follow-up studies explored frequency cues~\cite{frank2020frequency, bammey2023synthbuster} and handcrafted features~\cite{martin2023pixel} to enhance robustness. However, these models tend to degrade in performance when applied to images from unseen generative sources~\cite{gragnaniello2021gan}. Online adaptations~\cite{epstein2023online} attempt to mitigate this by continuously updating with new fakes, yet remain tightly coupled to evolving synthetic distributions, limiting long-term adaptability. Moreover, they produce outputs that, while able to separate classes, often lack statistical meaning and interpretability, as illustrated in Figure~\ref{fig:score_distribution}.

To address these limitations, recent work has shifted toward few-shot detectors that aim to reduce reliance on large labeled datasets. These approaches typically leverage powerful pre-trained encoders, such as Dino~\cite{oquab2023dinov2}, CLIP~\cite{radford2021clip, cozzolino2024raising, sha2023fake}, GAN~\cite{goodfellow2014generative, ojha2023towards}, Diffusion Models~\cite{chu2024fire, wang2023dire}  and adapt them using a small number of synthetic examples, fine-tuning or calibrating on limited fakes. These methods retain a degree of supervision. As a result, they require retraining to handle emerging models and remain tied to internal model-specific scores, limiting prediction interpretability and extensibility to future generators.
\captionsetup[subfigure]{font=small, skip=2pt, singlelinecheck=off, justification=centering}

\begin{figure}[!bt]
    \centering

    \begin{subfigure}[b]{0.48\linewidth}
        \includegraphics[width=\linewidth, keepaspectratio=false]{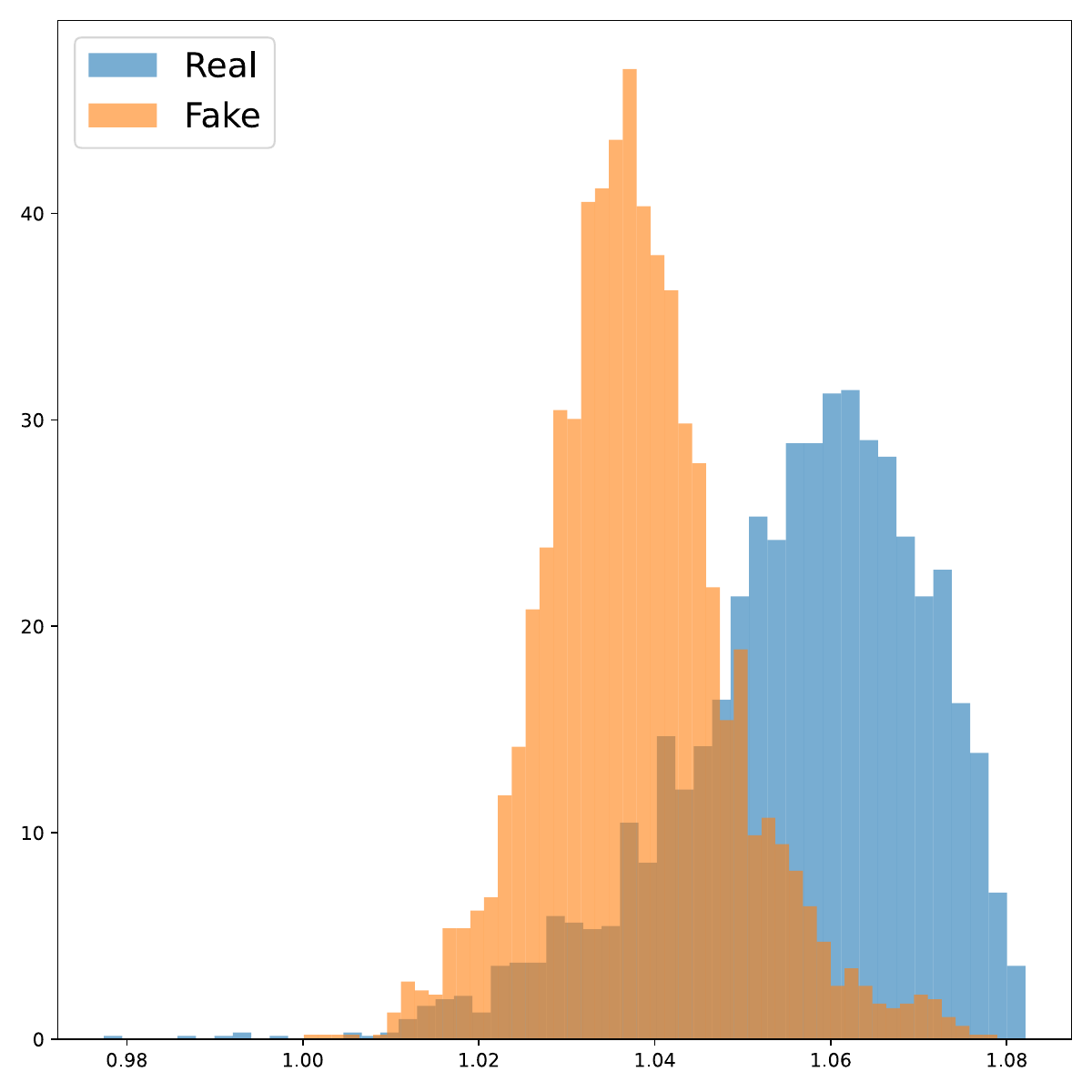}
        \caption{\strut Manifold Curvature (SDXL)}
    \end{subfigure}
    \hfill
    \begin{subfigure}[b]{0.48\linewidth}
        \includegraphics[width=\linewidth, keepaspectratio=false]{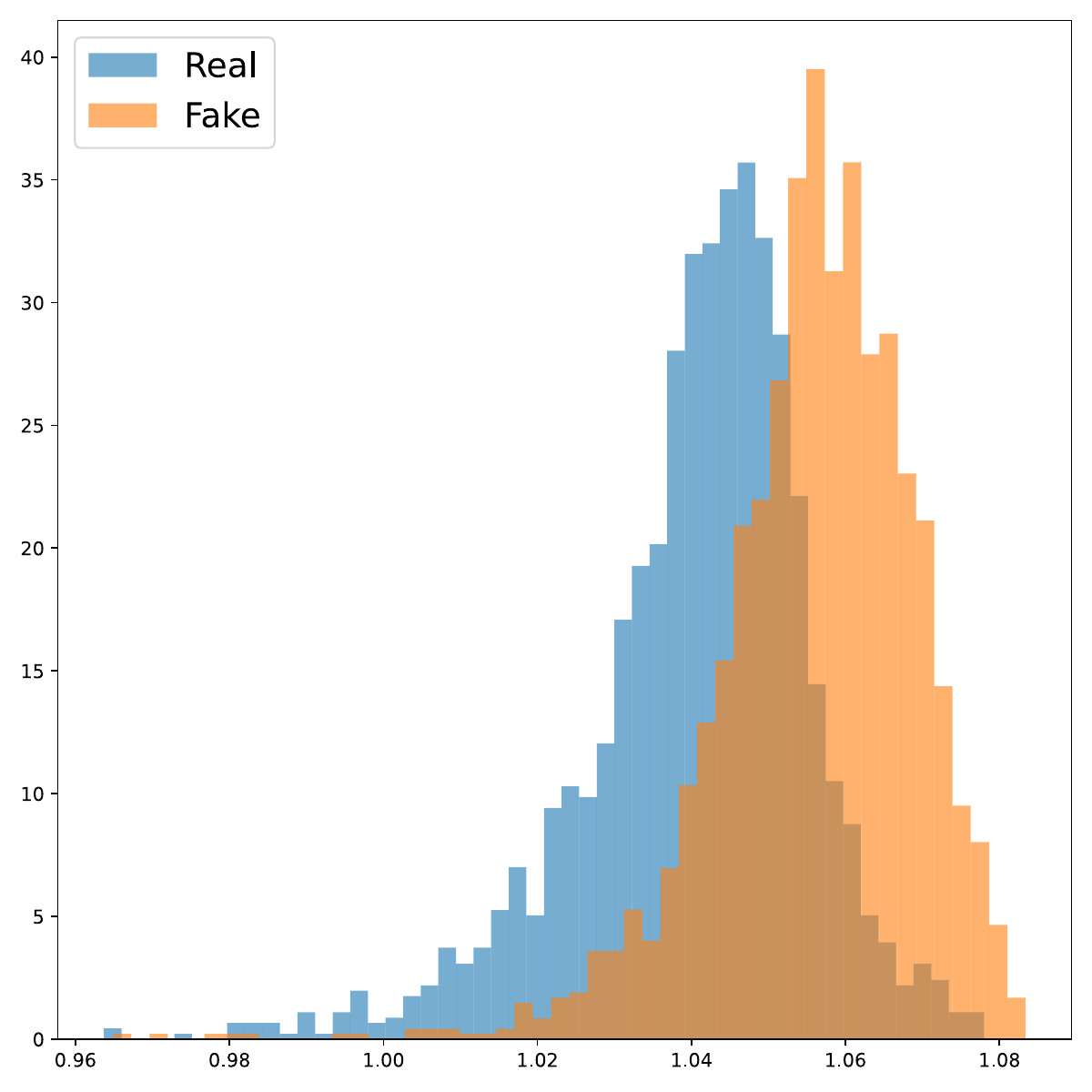}
        \caption{\strut Manifold Curvature (StyleGAN2)}
    \end{subfigure}

    \vspace{0.2cm}

    \begin{subfigure}[b]{0.48\linewidth}
        \includegraphics[width=\linewidth, keepaspectratio=false]{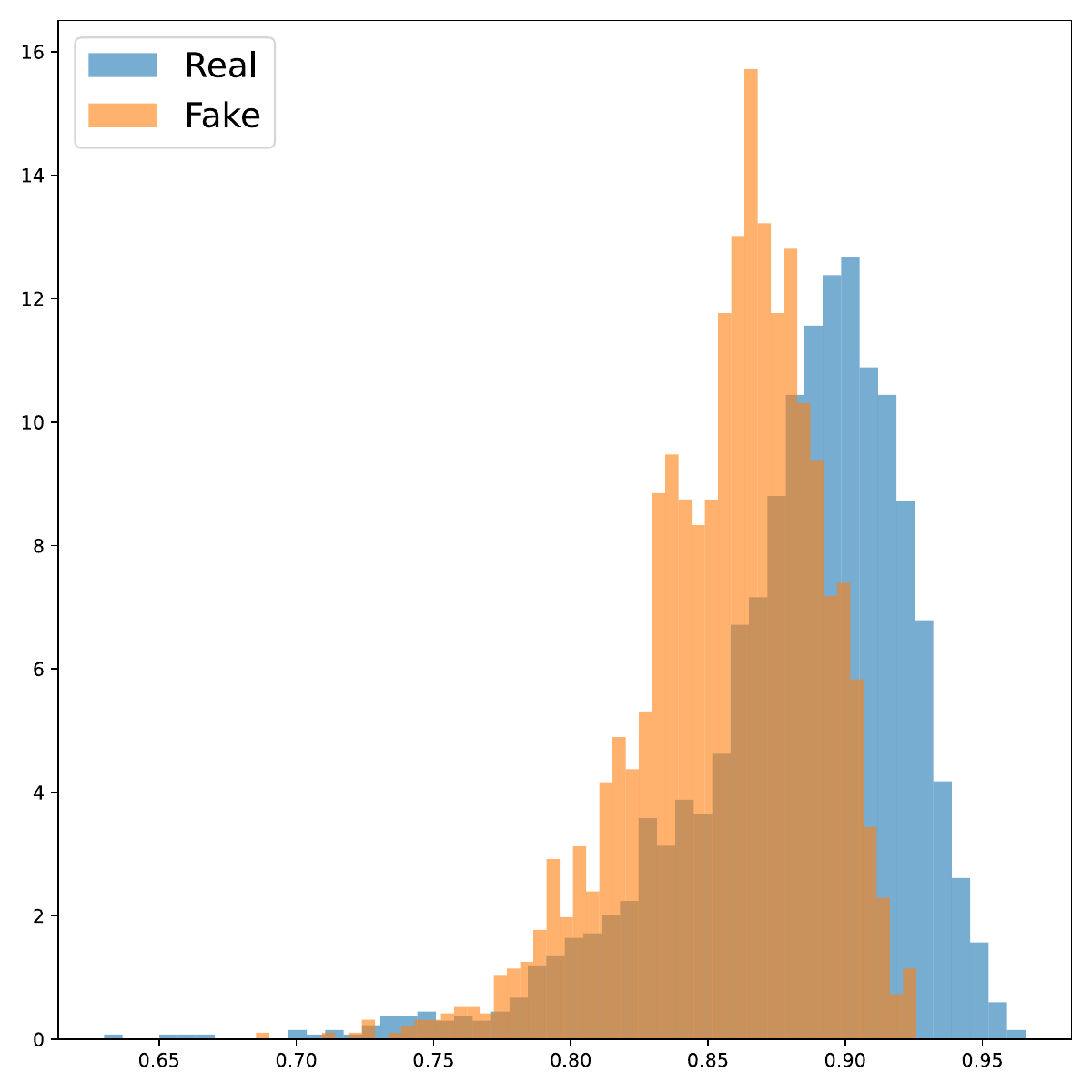}
        \caption{\strut Permutation-Based Features (StyleGAN2, $f$ = CLIP, $\lambda = 0.1$)}
    \end{subfigure}
    \hfill
    \begin{subfigure}[b]{0.48\linewidth}
        \includegraphics[width=\linewidth, keepaspectratio=false]{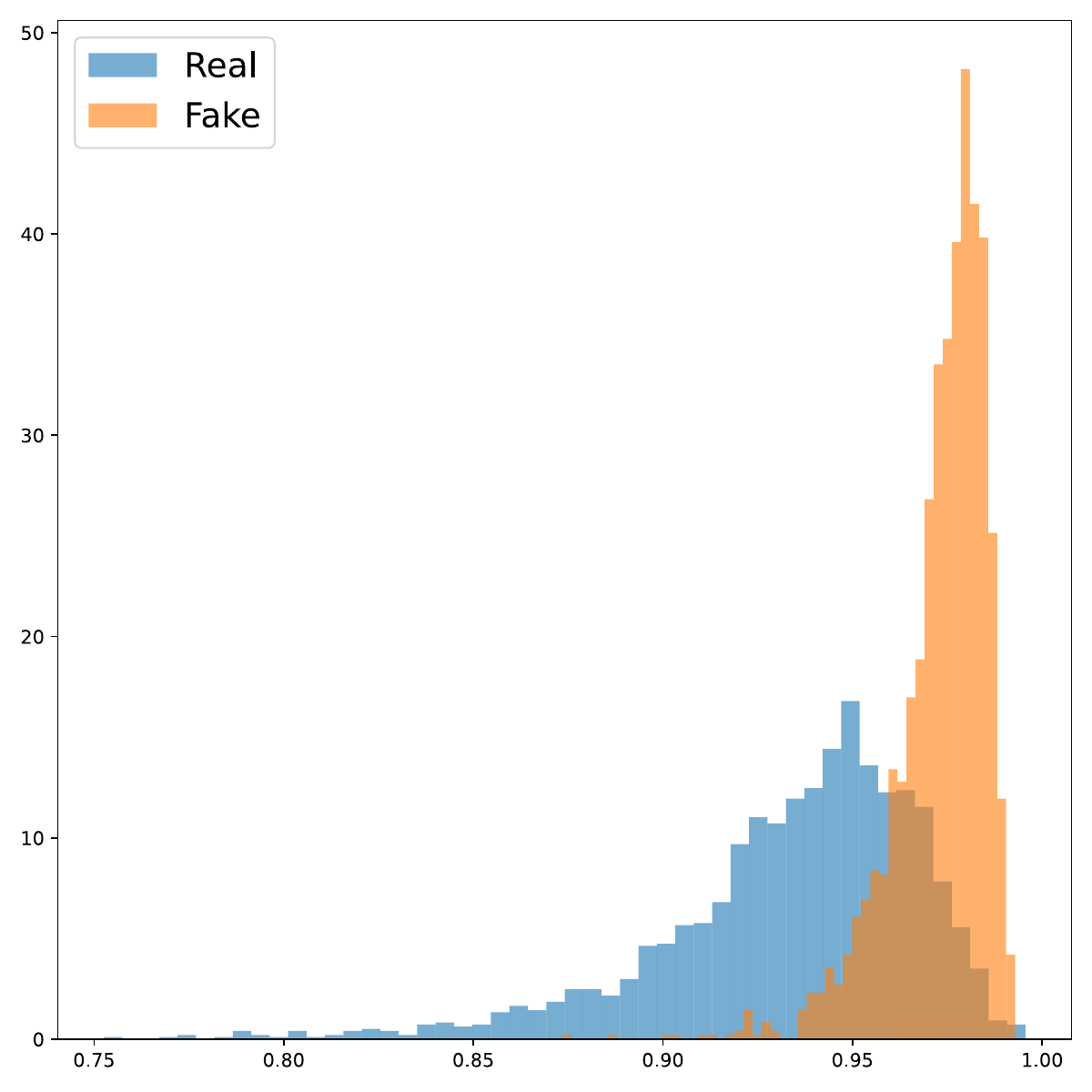}
        \caption{\strut Permutation-Based Features (StyleGAN2, $f$ = CLIP, $\lambda = 0.01$)}
    \end{subfigure}

    \caption{
    Illustration of the adaptability gap in existing training-free methods.
    Each row shows statistics from a different method (top: manifold curvature; bottom: permutation-based features), and each column corresponds to a different test condition.
    \textbf{Top:} The manifold curvature statistic shows opposite behavior across generators. In (a) SDXL, real images have higher curvature than real ones, while in (b) StyleGAN2, the pattern reverses.
    \textbf{Bottom:} The permutation-based feature statistic is sensitive to the perturbation strength \( \lambda \). In (c), real images yield higher scores than fakes, but in (d), the ordering flips.
    These shifts highlight that handcrafted statistics often rely on assumptions, such as fixed magnitude differences, that do not generalize across models or test settings.
    }

    \label{fig:statistics_distribution}
\end{figure}
A distinct class of methods avoids supervision by operating in a training-free, zero-shot regime. These approaches draw on intrinsic statistical properties of real images and evaluate test statistics without access to fake samples. Perturbation-based detectors, such as RIGID~\cite{he2024rigid}, assume that real images exhibit greater embedding stability under noise, using a similarity threshold to distinguish them from fakes. A related approach ~\cite{brokman2025manifold}, leverages the geometry of a generative model’s log-probability manifold, using curvature as a statistic. It assumes that generated images concentrate near local maxima, while real images lie in flatter regions. Similarly, reconstruction-based methods such as AEROBLADE~\cite{ricker2024aeroblade}, use latent diffusion model~\cite{rombach2022high} autoencoders to compute the reconstruction error, assuming that the generated images are better reconstructed than the real ones. These techniques rely on meaningful statistical measures that effectively distinguish real from generated images, offering a level of interpretability by grounding detection in well-defined statistics. However, they often assume a one-sided hypothesis test, which does not always hold in practice, as illustrated in Figure~\ref{fig:statistics_distribution}, which can reduce their ability to adapt to the behaviors of emerging generative models. In many cases, these methods are based on a single hand-crafted statistic and do not incorporate multiple signals. As a result, their generalization tends to be more effective within specific families of generative models.

Inspired by the effectiveness of recent zero-shot methods that leverage strong, model-independent statistics, we build on these insights to develop a unified, statistically rigorous framework. While prior approaches often rely on single handcrafted tests and assume fixed differences between real and fake statistics, typically calibrated through thresholds, our method extends these ideas without such assumptions. It integrates multiple detection signals using only real image distributions. By transforming scalar statistics into $p$-values and aggregating them with statistical methods, our approach provides interpretable outputs, principled error control, and improved adaptability to emerging generative models.

\section{Rationale}
\subsection{Detection as Statistical Hypothesis Testing}

We frame fake image detection as a statistical hypothesis multi-test~\cite{wasserman2004all}. For each scalar statistic \( s(x) \), we test whether an image \( x \) is plausible under the real image distribution.

Under the null hypothesis \( H_0 \colon x \sim \mathbb{P}_{\text{real}} \), we evaluate the extremeness of \( s(x) \) relative to a reference distribution estimated from real images \( \mathcal{D}_{\text{real}} = \{x_1, \dots, x_N\} \). The empirical cumulative distribution function (ECDF) is defined as:
\begin{equation}
\widehat{F}_N(t) = \frac{1}{N} \sum_{i=1}^N \mathbb{I}\{s(x_i) \leq t\}, \quad x_i \sim \mathbb{P}_{\text{real}}.
\label{eq:ecdf}
\end{equation}
Then, we compute a two-sided empirical $p$-value:
\begin{equation}
p(x) = 2 \cdot \min\left( \widehat{F}_N(s(x)),\, 1 - \widehat{F}_N(s(x)) \right).
\label{eq:twosided_pval}
\end{equation} 
This testing procedure is applied across multiple scalar statistics \( s_1(x), \dots, s_K(x) \), each designed to capture a distinct aspect of deviation from the real image distribution. In next sections, we describe how these individual $p$-values (as defined in Eq.~\eqref{eq:twosided_pval}) are aggregated into a single $p$-value.

\subsection{Validity Conditions}

The validity of the empirical $p$-value \( p(x) \) relies on the following assumptions about the real image dataset \( \mathcal{D}_{\text{real}} = \{x_1, \dots, x_N\} \) and the test statistic \( s(x) \):

\begin{enumerate}
    \item \textbf{i.i.d. sampling}: The reference samples \( x_1, \dots, x_N \) are drawn independently and identically from the true real image distribution: \( x_i \sim \mathbb{P}_{\text{real}} \).
    
    \item \textbf{Independent statistics}~The statistics \( s_1(X), \dots, s_K(X) \) used in aggregation are required to be independent under \( \mathbb{P}_{\text{real}} \). Rather than assuming this, the method enforces independence by selecting a subset that satisfies formal independence criteria as explained in subsequent sections.
    
    \item \textbf{distributional match}: A real test image \( x \) is also drawn from the same distribution: \( x \sim  \mathbb{P}_{\text{real}} \).
\end{enumerate}

Under these conditions, the $p$-value defined in Eq.~(2) is uniformly distributed on \([0,1]\) under the null hypothesis \( H_0 \) (see Appendix~\ref{appendix:proofs}, Lemma 1).

\begin{figure*}[!ht]
    \centering
    \includegraphics[width=0.96\textwidth]{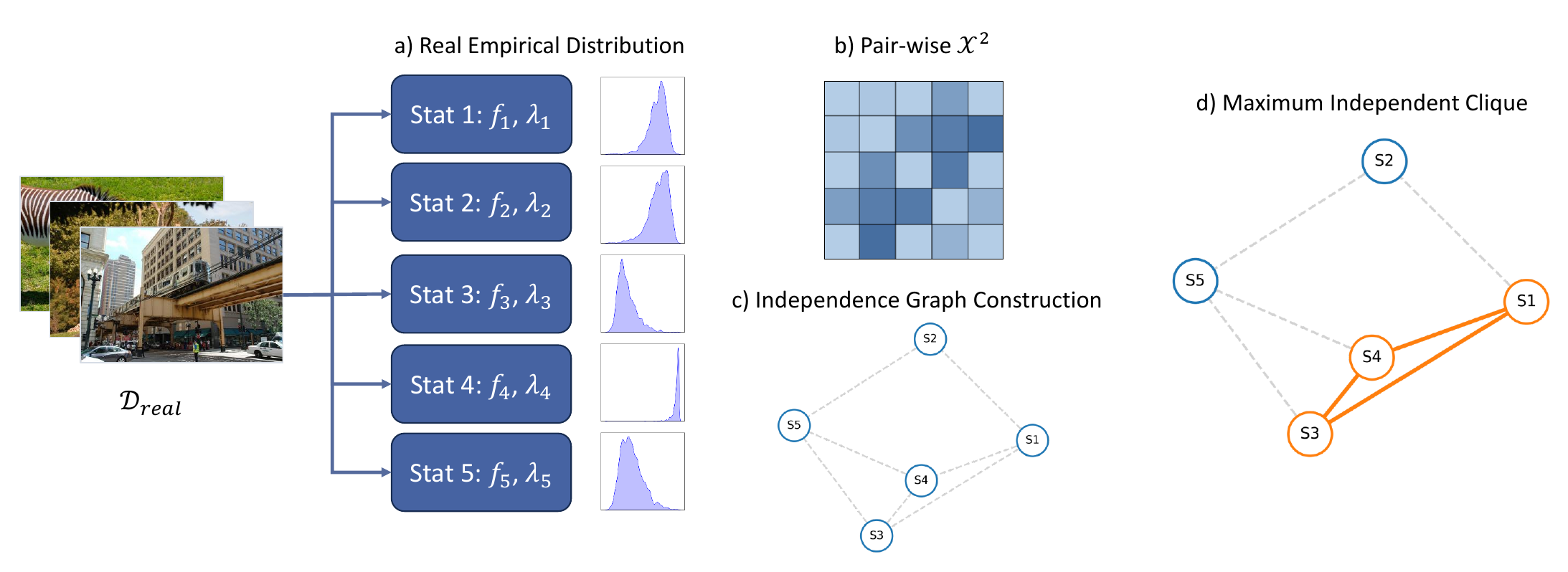}
    \caption{
        Overview of the null distribution modeling phase. (a) Real images from the reference dataset \( \mathcal{D}_{\text{real}} \) are processed using multiple detector configurations \( (f_j, \lambda_k) \), producing scalar statistics whose empirical distributions are estimated and stored as ECDFs. (b) Pairwise statistical dependence among statistics is assessed via \(\chi^2\) tests over real samples. (c) The resulting relationships define an independence graph, where edges indicate accepted independence. (d) A maximal clique is extracted and regularized via a uniformity constraint to select a subset \( \mathcal{I} \subseteq \mathcal{S} \) of independent statistics.
    }
    \label{fig:calibration}
\end{figure*}
\section{Proposed Method}
We propose a training-free detection method that tests whether an image is drawn from the distribution of real images, based on a rigorous statistical framework. The method consists of two phases. In the \textit{null distribution modeling} phase, we compute a diverse collection of scalar statistics over real images and estimate their ECDFs. Then we find a subset of independent statistics that allow valid multi-test aggregation under the null hypothesis (see Fig.~\ref{fig:calibration}). In the \textit{inference} phase, selected statistics are computed and mapped to \( p \)-values via stored ECDFs, then aggregated into a single interpretable \( p \)-value under the null distribution.

Each step of the null distribution modeling and detection pipeline is described in detail below.

\subsection{Null Distribution Modeling Phase}
\label{sec:null_modeling}

This phase has two stages: \textit{statistic extraction and empirical modeling} and \textit{independent subset selection}. Given a reference dataset \( \mathcal{D}_{\text{real}} \), it produces (1) set of stored ECDFs and (2) subset of independent statistics, enabling statistically valid \( p \)-value aggregation during detection.

\newcounter{stage}
\newcounter{step}[stage]

\renewcommand{\thestage}{\arabic{stage}}
\renewcommand{\thestep}{\thestage.\arabic{step}}

\stepcounter{stage}
\subsubsection*{Stage \thestage: Statistic Extraction and Empirical Modeling}
\label{sec:statistic_modeling}

In this stage, we compute a diverse set of scalar statistics \( \mathcal{S}(x) \) over real images \( x \in \mathcal{D}_{\text{real}} \), These statistics are computed using a broad set of feature extractors. For each statistic in \( \mathcal{S} \), we compute and store its ECDF, forming the reference needed for hypothesis testing during detection.

\stepcounter{step}\noindent\textbf{Step \thestep: Multi-Detector Processing}
\label{sec:detectors}

Each image \( x \) is processed by a set of frozen feature extractors \( \mathcal{F} = \{f_1, \dots, f_m\} \), where each \( f_j \colon \mathbb{R}^{h \times w \times C} \rightarrow \mathbb{R}^d \) maps an image to a high-dimensional embedding.

Inspired by RIGID~\cite{he2024rigid}, we assess the stability of these embeddings by applying small Gaussian perturbations to the input and measuring the change in the resulting features. Specifically, we define additive noise \( \delta \sim \mathcal{N}(0, I) \) and select a perturbation strength \( \lambda_k \in \{\lambda_1, \dots, \lambda_n\} \). The image score is then computed as the cosine similarity between clean and perturbed features:
\begin{equation}
s_{j,k}(x) = \text{sim}\left(f_j(x), f_j(x + \lambda_k \delta)\right).
\label{rigid_statistic}
\end{equation}

This score reflects the robustness of the embedding under noise: real images typically produce more stable features, leading to higher similarity scores. However, this behavior is not universally consistent across encoders or noise level. To increase robustness and sensitivity, we evaluate a wide set of detector configurations, each defined by a pair \( (f_j, \lambda_k) \) combining a vision backbone and a perturbation level.

The complete set of scalar statistics extracted from image \( x \) is defined as:
\[
\mathcal{S}(x) = \left\{ s_{j,k}(x) \right\}_{j \leq m,\; k \leq n}.
\]

The modular construction of \( \mathcal{S}(x) \) allows new detectors to be easily incorporated, providing a clear path for future extension and adaptability.

\stepcounter{step}\noindent\textbf{Step \thestep: Empirical Two-Sided $p$-Value Estimation}
\label{sec:ecdf}

For each scalar statistic in \( \mathcal{S}(x) \), we evaluate its extremeness relative to the distribution of the same statistic computed over a reference set of real images.

For each configuration independently, we construct an ECDF as defined in Eq.~\eqref{eq:ecdf}.

Given image \( x \), we compute a two-sided \( p \)-value for each statistic using Eq.~\eqref{eq:twosided_pval}, resulting in a \( p \)-value vector:
\[
\mathbf{p}(x) = \left[p_{j,k}(x)\right]_{(j,k)}.
\]
The two-sided formulation avoids assuming a fixed direction of deviation and accommodates diverse behaviors across detectors.

The ECDFs \( \widehat{F}_s \) are estimated using real images only, which we treat as a representative sample drawn from the real population, allowing the method to remain adaptable as generative techniques evolve.

\stepcounter{stage}
\subsubsection*{Stage \thestage: Independent Subset Selection}
\label{sec:independence_selection}
In this stage, we select a subset of scalar statistics that are independent under the null hypothesis. This enables statistically valid aggregation of their corresponding \( p \)-values. The selection is performed by testing pairwise independence among statistics, constructing an independence graph, and finding a maximal clique under a uniformity constraint.

\stepcounter{step}\noindent\textbf{Step \thestep: Pairwise Dependence Testing} 

Stacking the vectors \( \mathbf{p}(x_n) \) for all \( x_n \in \mathcal{D}_{\text{real}} \), we form an \( N \times T \) matrix of \( p \)-values, where \( T = |\mathcal{S}| \).

To assess mutual dependence between statistics, we apply a pairwise \(\chi^2\) test to all \( T(T-1)/2 \) pairs of statistics, testing whether the joint distribution of \( p \)-values over real images deviates from independence.

\stepcounter{step}\noindent\textbf{Step \thestep: Independence Graph Construction}

We construct an undirected graph \( G = (\mathcal{V}, \mathcal{E}) \), where each node \( v_i \in \mathcal{V} \) represents a statistic \( s_i \in \mathcal{S} \). For each pair \( (s_i, s_j) \), dependence is assessed via the $\chi^2$ statistic and quantified using Cramér’s \( V \), which avoids the instability of $\chi^2$ $p$-values in large samples~\cite{Lin2013ResearchC}. An edge is added if the association is weak, i.e.,
\[
(v_i, v_j) \in \mathcal{E} \quad \text{iff} \quad V(s_i, s_j) \leq \mathcal{V}_{\text{$\chi^2$}}.
\]

where \(V_{\chi^2}\) denotes the Cramér’s \(V\) threshold. This graph captures the empirical structure of pairwise associations sufficiently weak to approximate independence.

\stepcounter{step}\noindent\textbf{Step \thestep: Maximum Clique Enumeration} 

We extract a maximal clique~\cite{maxcliquealgo} from the independence graph, corresponding to the largest subset \( \mathcal{I} \subseteq \mathcal{S} \) of pairwise independent statistics. 
However, pairwise independence alone does not guarantee joint independence, which is required for valid multi-test aggregation. (details on aggregation methods in Section~4.2).

To address this, we select \( \mathcal{I} \) by verifying that the aggregated \( p \)-values follow a uniform distribution under the null hypothesis. To mitigate large-sample artifacts analogous to those affecting $\chi^2$ $p$-values, we apply the Kolmogorov--Smirnov test to a representative subsample of \( N_{\text{KS}} \) real images:
\[
p\text{-value}_{\text{KS}} \geq \alpha_{\text{KS}}.
\]

This ensures that the selected set supports valid downstream aggregation under the null hypothesis. 
Aggregation methods such as Stouffer’s test or the min-\( p \)-value method (described in the next section) quantify the joint deviation from the null across statistics. 

In practice, detection performance remains an important consideration. Although the set of statistics is chosen based on independence and null alignment, encoders such as DINOv2 and CLIP are known to produce particularly strong discriminative features. Thus, when multiple maximal cliques satisfy these conditions, we prioritize those containing these encoders to improve effectiveness.

\begin{figure*}[!ht]
    \centering
    \includegraphics[width=0.96\textwidth]{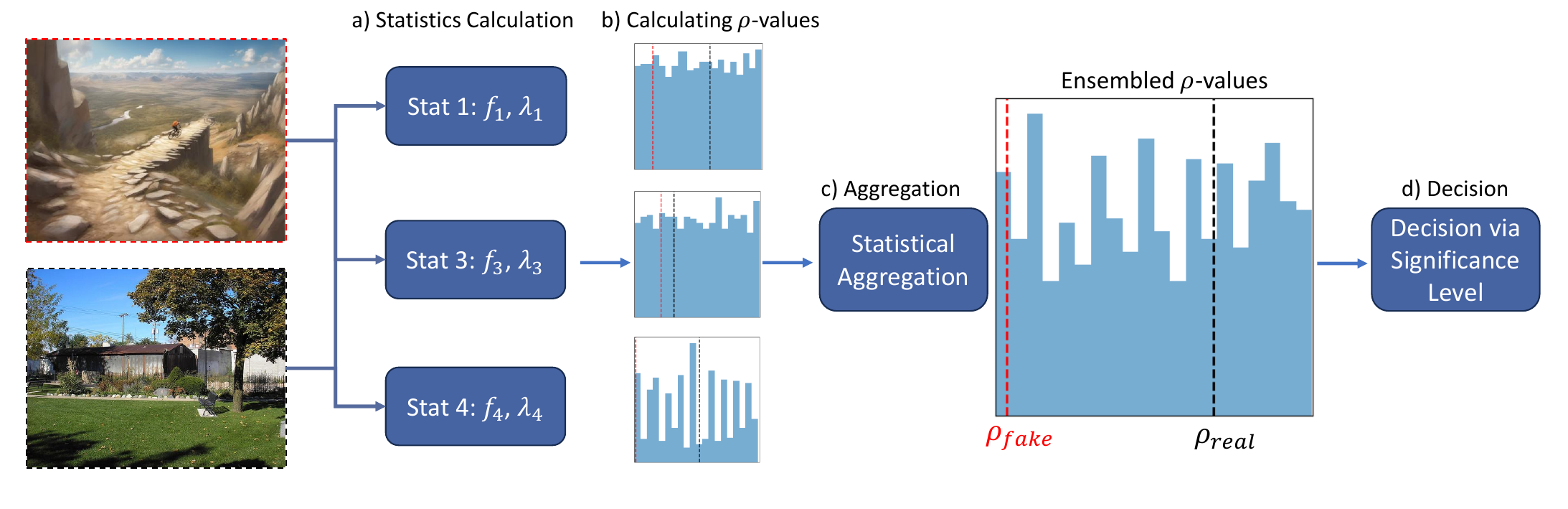}
    \caption{
        Overview of the inference phase. (a) A candidate image is processed using only the subset of detector configurations corresponding to the selected statistics \( \mathcal{I} \). (b) Each resulting statistic is mapped to a two-sided \( p \)-value using the stored ECDFs. (c) The set of \( p \)-values is aggregated using a statistical aggregation method, such as Stouffer’s test. (d) The final decision is taken by comparing the unified \( p \)-value to a predefined significance level.
    }
    \label{fig:detection}
\end{figure*}
\subsection{Inference Phase}

Given a candidate image \( x \), we compute only the subset of scalar statistics \( \mathcal{I} \subseteq \mathcal{S} \) that were selected during the null distribution modeling phase as independent (Step.~2.3), as illustrated in the detection pipeline (Fig.~\ref{fig:detection}). To do so, we apply multi-detector processing (Step.~1.1), restricted to the relevant configurations that define \( \mathcal{I} \).

Each statistic \( s \in \mathcal{I} \) is then mapped to a two-sided \( p \)-value using the corresponding ECDF \( \widehat{F}_s \) estimated from real images (Step.~1.2). These individual \( p \)-values quantify how extreme each selected statistic is under the null hypothesis.

To summarize this evidence into a single interpretable score, we apply one of two statistical aggregation methods, both of which assume independence among the selected statistics.

\paragraph{Stouffer’s Test:} This method transforms each \( p \)-value into a standard normal Z-score:
\begin{equation}
z_i = \Phi^{-1}(p_i)
\end{equation}
\begin{equation}
Z = \frac{1}{\sqrt{K}} \sum_{i=1}^K z_i, \quad P_{\text{Stouffer}} = \Phi(Z)
\label{eq:stouffer_z}
\end{equation}
where \( \Phi \) is the standard normal CDF and \( K = |\mathcal{I}| \) is the number of aggregated statistics. Under the null, \( P_{\text{Stouffer}} \sim \mathcal{U}[0, 1] \) (see Appendix~\ref{appendix:proofs}, Lemma~2). This method is effective when several statistics show moderate deviations from the null that, while individually weak, become significant when combined.

\paragraph{Minimum \( p \)-Value:} This method emphasizes the strongest individual evidence:
\begin{equation}
P_{\min} = \min_{i} p_i, \quad F_{P_{\min}}(t) = 1 - (1 - t)^K
\label{eq:min_pval}
\end{equation}
where \( P_{\min} \) is the minimum of \( K \) independent \( p \)-values, and \( F_{P_{\min}} \) is its CDF. Under the null hypothesis, and assuming \( p \)-values are computed as in Eq.~\eqref{eq:twosided_pval}, the resulting test remains valid with uniform distribution (see Appendix~\ref{appendix:proofs}, Lemma~3). This method is well suited when only a small subset of detectors is expected to distinguish fakes.

The resulting unified \( p \)-value provides a statistically valid and interpretable measure of authenticity, quantifying the deviation of the candidate image \( x \) from the distribution of real images. \\

The complete procedures for null modeling and inference are detailed in Appendix~\ref{appendix:pseudocode}, while a comprehensive study of runtime and memory usage is provided in Appendix~\ref{appendix:runtime_memory}.

\section{Experiments}

\paragraph{Datasets}
We evaluate our method on several large-scale benchmark datasets that collectively span a broad spectrum of generative models and content domains. The CNNSpot dataset~\cite{wang2020cnn} consists of real and synthetic images across 20 LSUN~\cite{yu2015lsun} categories, primarily generated using early convolutional and GAN-based models. The Universal Fake Detect dataset~\cite{ojha2023towards} expands this setup to include more recent latent diffusion architectures. To further assess robustness on high-fidelity diffusion-based content, we incorporate a Stable Diffusion Face dataset that provides high-quality generated images from SDXL and SDv2~\cite{stable_diffusion_face_dataset}. Finally, to evaluate performance on challenging real-world generative systems, we include the Synthbuster~\cite{bammey2023synthbuster} and GenImage~\cite{zhu2023genimage} datasets.

Across these datasets, our aggregated dataset comprises a total of 187K images (93K real and 94K fake), approximately balanced between real and generated content. This large-scale and diverse benchmark enables evaluation of detection performance across multiple generation paradigms. A complete list of generative models is provided in Appendix~\ref{appendix:generators}, and the exact dataset splits and reproducibility details are described in Appendix~\ref{appendix:reproducibility}.

\paragraph{Baselines}
We compare our method against three recent training-free, approaches that represent complementary statistical detection strategies.
RIGID measures embedding stability under perturbation and serves both as a strong standalone baseline and as a foundational component within our framework.
AEROBLADE evaluates reconstruction error using latent autoencoders, while
ManifoldBias~\cite{brokman2025manifold} quantifies statistical curvature in the latent space of pre-trained diffusion models.

\paragraph{Implementation Details}
Our method relies on frozen visual encoders and Gaussian perturbations. All detectors operate at image resolutions of 512x512. The specific encoders and perturbation strengths used in our experiments, including hyper-parameters are summarized in Appendix~\ref{appendix:detector_config} and Appendix~\ref{appendix:hyperparams} respectively.

To ensure a statistically valid evaluation, we explicitly partition the real images into two disjoint subsets: 30\% is allocated to the null distribution modeling phase (i.e., ECDF estimation), while the remaining 70\% is held out for evaluation. This split ensures that no test-time inference image is used during calibration. Fake images are evaluated only against the held-out portion of real images. All detection methods, including ours and the baselines, are evaluated under the same protocol using shared real and fake subsets for each generator.

At the time of writing, the official implementation of RIGID is not publicly available. We implemented the method based on the description in the original paper, using its best-reported configuration: a DINOv2 backbone with a perturbation strength of 0.05. Due to variability in reported results across different works, we will release our full implementation, which includes both the original RIGID setup and extended variants with alternative feature extractors and perturbation levels.

\paragraph{Metrics}
We report two threshold-free metrics to evaluate detection performance: \textit{Area Under the ROC Curve (AUC)} and \textit{Average Precision (AP)}, each computed per generative model. Unlike methods that require selecting a fixed threshold, our approach produces $p$-values from hypothesis testing. Threshold-based metrics like accuracy are thus not directly comparable and may not reflect statistical confidence. AUC and AP provide a more appropriate view of performance across the decision range.

\subsection{Interpretability Does Not Come at the Price of Performance}
A central goal of our method is to provide interpretable and
reliable detection. Interpretability is a direct result of returning a $p$-value for each inference image $x$, which is a value with clear statistical interpretation, rather than an ambiguous "realness score". Crucially, we demonstrate that despite this design choice, our method achieves performance on par with state-of-the-art training-free detectors.

\begin{table}[htb]
    \centering
    \small
    \caption{Average AUC and AP (with standard deviation) across all generators splits.}
    \label{tab:auc_summary}
    \begin{tabular*}{\linewidth}{@{\extracolsep{\fill}}lcc}
    \hline
    Model & AUC & AP \\
    \hline
    Manifold Bias   & 0.761 $\pm$ 0.179  & 0.753 $\pm$ 0.169 \\
    RIGID           & 0.769 $\pm$ 0.194  & 0.765 $\pm$ 0.189 \\
    AEROBLADE       & 0.697 $\pm$ 0.161  & 0.697 $\pm$ 0.163 \\
    Ours (Stouffer) & 0.756 $\pm$ 0.135  & 0.743 $\pm$ 0.133 \\
    Ours (Min-$p$)  & 0.775 $\pm$ 0.126  & 0.756 $\pm$ 0.119 \\
    \hline
    \end{tabular*}
\end{table}

\noindent\textit{Note.} Each split is balanced between real and fake samples using stratified sampling to ensure fair comparison across methods.

As shown in Table~\ref{tab:auc_summary}, our method achieves competitive AUC and AP compared to state-of-the-art training-free baselines. While Manifold Bias and RIGID obtain slightly higher peak scores, our approach offers comparable performance with substantially lower variance across generators, reflecting more consistent behavior. Importantly, this competitiveness comes without sacrificing interpretability

\begin{figure}[htbp]
    \centering
    \includegraphics[width=0.48\textwidth]{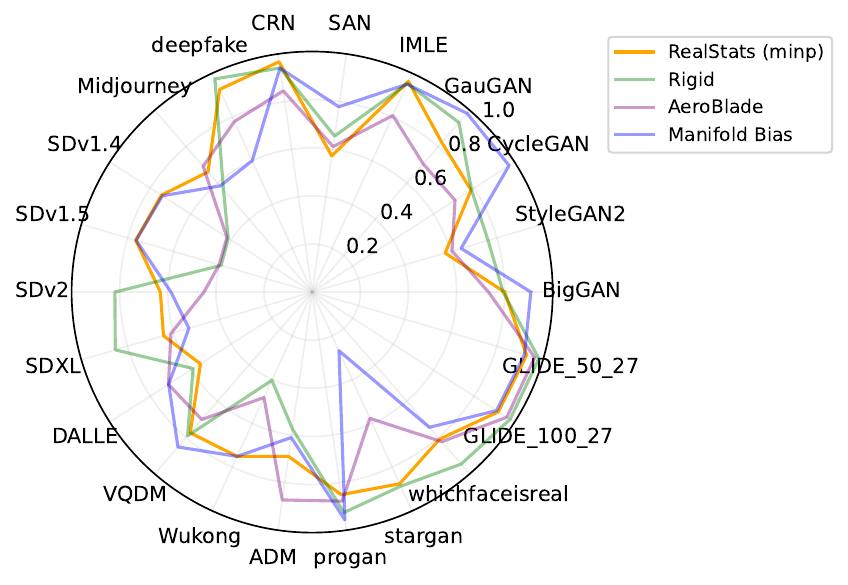}
    \caption{
    Per-generator AUC comparison across methods shown in radar format.
    }
    \label{fig:auc_per_generator}
\end{figure}

Empirically, different methods peak on some generators but drop on others, showing inconsistent stability (see Figure~\ref{fig:auc_per_generator}). Manifold Bias, for instance, scores highly on GauGAN but drops on StarGAN and SDv2. RIGID performs well on SDXL but struggles on SDv1.4, while AeroBlade is strong on ADM but weak on SDv1.5. Our method shows some weaknesses on datasets like CycleGAN, yet overall maintains more balanced results, supported by its multi-RIGID variants that prevent collapse when a single statistic fails. 
Complete AUC and AP values are provided in Appendix~\ref{appendix:full_auc_ap}.

\subsection{Adaptability in Action: Improving Performance on Challenging Generators}

While our method performs on par with state-of-the-art training-free detectors overall, we observe weaker results on certain generators where \textit{ManifoldBias} excels, including \textbf{GauGAN}, \textbf{CycleGAN}, and \textbf{SAN} (see Figure~\ref{fig:auc_per_generator}). To demonstrate adaptability, we revisit these cases and integrate the \textit{ManifoldBias} statistic under the same experimental setup.

\begin{figure}[ht]
    \centering
    \includegraphics[width=0.94\linewidth]{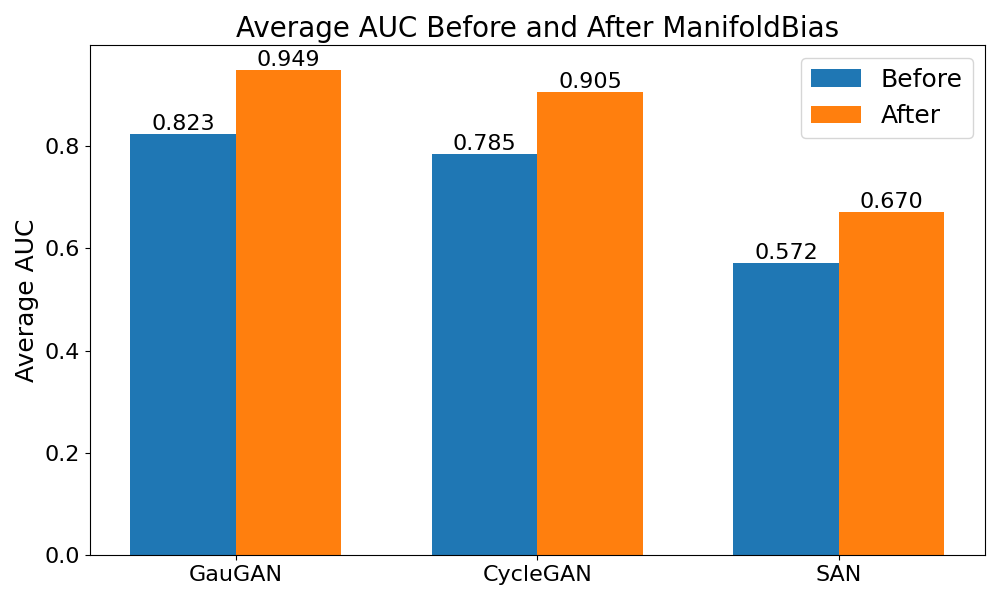}
    \caption{
        Average AUC of the Min-$p$ ensemble before and after incorporating \textit{ManifoldBias} on GauGAN, CycleGAN, and SAN.
    }
    \label{fig:auc_improvement_barplot}
\end{figure}

As shown in Figure~\ref{fig:auc_improvement_barplot}, the ensemble gains clear improvements on these generators, closing the gap with baselines by leveraging the strength of \textit{ManifoldBias}. 

\begin{figure}[htbp]
    \centering
    \begin{subfigure}[b]{0.49\linewidth}
        \includegraphics[width=\linewidth]{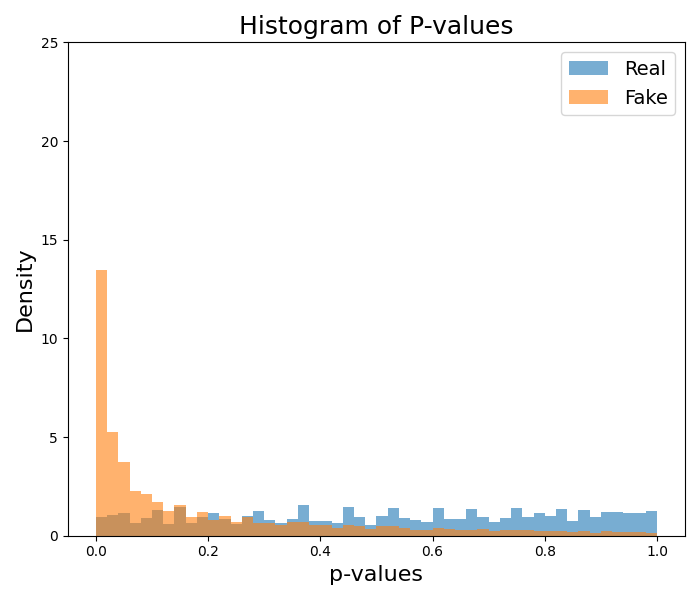}
        \caption{\strut Entire Dataset, Min-$p$ ensemble (before)}
    \end{subfigure}
    \hfill
    \begin{subfigure}[b]{0.48\linewidth}
        \includegraphics[width=\linewidth]{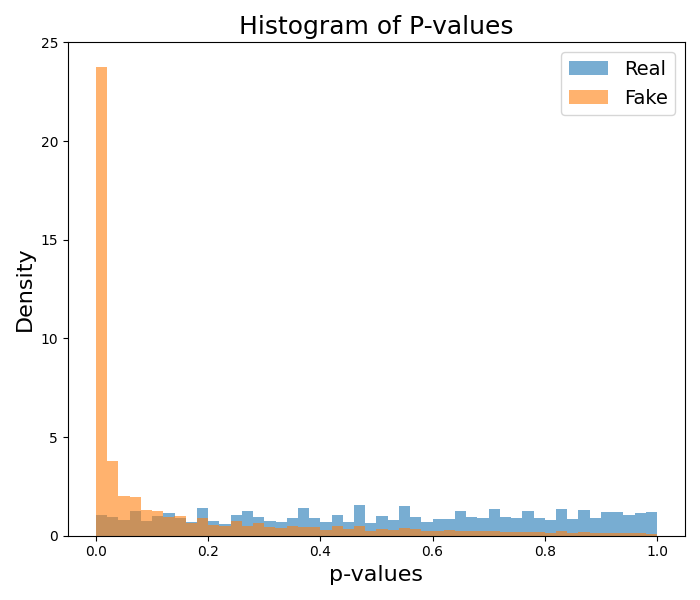}
        \caption{\strut Entire Dataset, Min-$p$ ensemble (after)}
    \end{subfigure}
    \caption{
        Example of improved \( p \)-value separation with \textit{ManifoldBias} in the Min-$p$ ensemble. Before (left), real and fake distributions overlap considerably; after (right), fake samples shift toward zero while real remain uniform.
    }
    \label{fig:adaptability_bias}
\end{figure}

To highlight the broader impact, the benefit extends to our full test benchmark of 160K images, where incorporating \textit{ManifoldBias} improves \( p \)-value separation (Figure~\ref{fig:adaptability_bias}) and raises overall AUC scores. These results demonstrate the modularity and adaptability of our framework, showing it can respond to evolving generative content by selectively integrating new, independent statistics when needed.

\subsection{Interpretability Qualititive Comparison}

Appendix~\ref{appendix:interpretability_qualitative} provides a qualitative comparison illustrating the interpretability of our method, where $p$-value patterns reveal how image deviations from the reference distribution manifest in a transparent and meaningful way.

\subsection{Fast, Scalable, and Memory-Efficient}

Our method is designed for high-throughput inference and scales efficiently with GPU parallelism. Runtime improves with more workers, while memory remains moderate even with many statistics and large batches. In addition, we demonstrate that the computational overhead of the independence-testing stage (Phase~1) is negligible relative to the forward-pass cost.

Unlike training-free methods built on heavy autoencoders (e.g., Stable Diffusion), it achieves faster inference and lower memory use. The approach runs on single- or multi-GPU setups with minimal overhead, making it practical for large-scale deployment.

Full runtime, scalability, and memory analyses are provided in Appendix~\ref{appendix:runtime_memory}.

\subsection{Effect of Reference Distribution Misalignment}

The empirical null modeling procedure relies on the assumption that the set of real images used to construct the ECDFs provides a representative sample of the real-image population encountered at inference time. Two distinct aspects can challenge this assumption. 

The first is finite sampling, where the empirical ECDF may not perfectly approximate the true CDF of the real-image population since the reference set is finite. In this case, the ECDF becomes a biased or coarse estimate: the outputs still provide useful relative ranking for distinguishing real from fake images, but the formal probabilistic meaning of a $p$-value under the null is no longer guaranteed. Because real images are generally easy to obtain, this limitation can often be mitigated by modestly enlarging or diversifying the reference set, and a practical way to assess representativeness is to compare the empirical distribution of $p$-values on a validation set to the expected uniform distribution. 

The second is distributional shift, where the test-time real distribution differs structurally or semantically from the reference domain, even if the reference set is large. Such shifts occur, for example, when content categories are heavily unbalanced, resolution changes, or new semantic attributes appear. In such cases, the $p$-values deviate from the expected uniform behavior, meaning they no longer function as true $p$-values in the classical probabilistic sense.
\begin{figure}[htbp]
    \centering
    \includegraphics[width=0.49\textwidth]{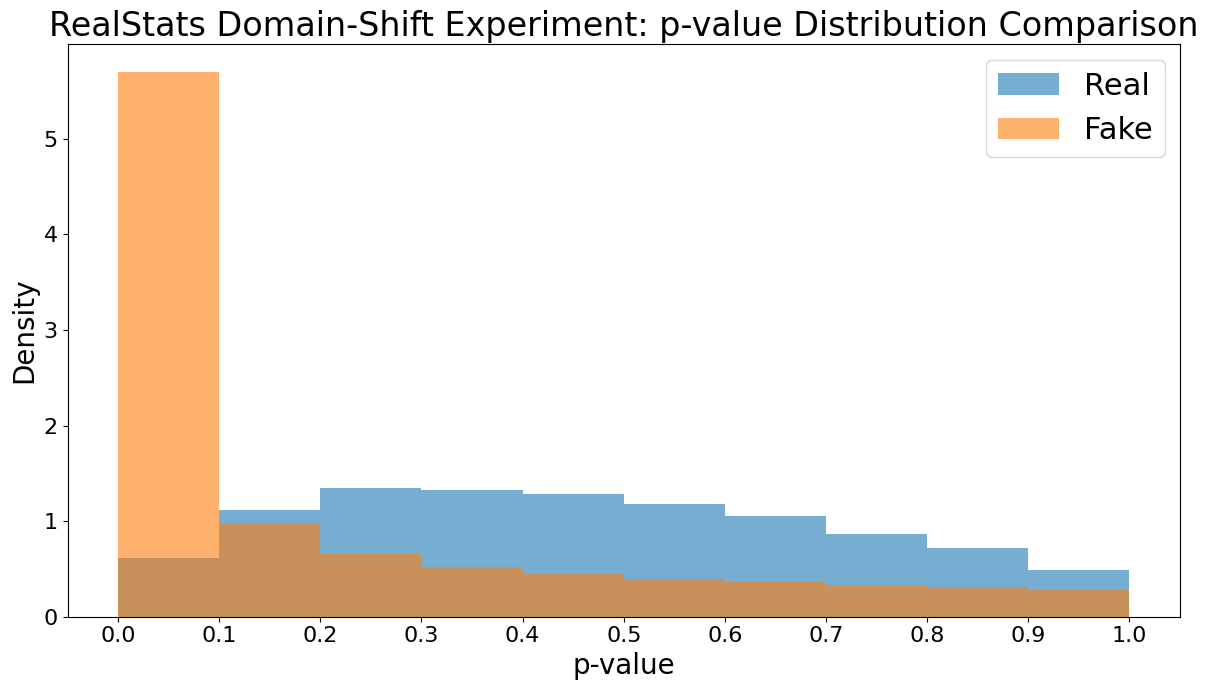}
    \caption{
        $p$-value density distributions under reference-test domain mismatch.
        Real samples (blue) deviate from uniform behaviour due to calibration mismatch, while fake samples (orange) remain concentrated near zero, demonstrating preserved discriminative signal.
    }
    \label{fig:distribution_shift_hist}
\end{figure}
To examine this effect, we designed a controlled domain-shift experiment in which the reference ECDF was constructed from only 1k FFHQ~\cite{karras2020analyzing} faces ($<4\%$ of the test size), while inference was performed on 30k high-resolution CelebA~\cite{liu2015faceattributes} faces. In this setup, faces in the test domain are abundant and diverse, whereas in the reference set they represent only a small and less varied subset. Moreover, CelebA differs from FFHQ both in structural properties (in particular, higher resolution than the vast majority of reference samples) and in overall domain composition (e.g., pose and attribute distribution), making it an appropriate setting for examining how \textsc{RealStats} behaves when the reference distribution captures only a limited portion of the true test domain.

Figure~\ref{fig:distribution_shift_hist} visualizes the resulting $p$-value densities for real and fake samples under this mismatch. Real samples deviate from uniformity, indicating calibration misalignment, while fake samples remain concentrated near zero, showing preserved discriminative signal.

Even in this scenario, \textsc{RealStats} maintains strong real-fake separation (AUC = 0.79, AP = 0.845), outperforming a perturbation-based heuristic evaluated under identical conditions (AUC = 0.70, AP = 0.74). This indicates that while the misalignment violates the strict probabilistic meaning of the $p$-values, the aggregated scores remain highly discriminative.

\subsection{Robustness Under Image Corruptions}

We evaluate robustness to JPEG compression and Gaussian blur at test time without altering the reference distribution or evaluation setup. With the Min-$p$ ensemble, blur keeps performance stable or slightly improved, while JPEG compression causes a moderate drop (5\% AUC, 6.4\% AP) but does not affect the validity of $p$-values. Overall, the framework remains resilient to these corruptions. 

Additional results and visualizations are provided in Appendix~\ref{appendix:robustness_corruptions}.



\section{Limitations}

Our framework is subject to two main limitations.  
First, the performance of our method depends on the selected statistic clique. In some cases, excluding correlated yet informative detectors may reduce separability. To mitigate this, we prioritize valid cliques that include strong and diverse detectors such as DINOv2 and CLIP.

Second, the validity of the resulting $p$-values depends on the quality of the reference distribution. As discussed earlier, both finite sampling and distributional shift can affect the representativeness of the empirical null model. Reliability can be assessed by examining the $p$-value distribution on a validation set and comparing it with the expected uniform pattern. In practice, misalignment is often manageable, as real data is typically easy to obtain.

\section{Conclusion}
We presented a statistical framework for fake image detection focused on two main properties: interpretability and adaptability. Through extensive experiments, we showed that our method achieves competitive performance with state-of-the-art training-free detectors, while remaining robust, scalable, and modular.

\clearpage

\bibliography{aistats2026}    

\clearpage
\appendix



\onecolumn

\aistatstitle{Supplementary Materials}

\appendix
\section*{Appendix}
This appendix contains supplementary material that supports the methods and findings in the main paper. It includes:
\begin{itemize}
    \item A. Algorithmic Details: Pseudocode for the two phases of the framework, detector configurations, and hyperparameter settings.
    
    \item B. Statistical Validity: Lemmas justifying the validity of the proposed $p$-values construction and aggregation.

    \item C.  List of Generative Models: Comprehensive list of generative models used in evaluation.
    
    \item D.  Full Experimental Results: Detailed baseline comparison table, runtime and memory analysis, robustness evaluations, and qualitative visualization of interpretability.

    \item E. Dataset and Code: Description of released data splits, code and setup for reproducibility.
\end{itemize}

\section{Algorithmic Details}
This section provides a detailed algorithmic overview of the phases and the components that constitute our framework.

\subsection{Detectors Configurations}
\label{appendix:detector_config}

\begin{table}[htbp]
\centering
\caption{Configuration of the statistics set used across experiments. Each row specifies the feature extractor and the perturbation strengths \( \lambda \) applied during statistic extraction.}
\begin{tabular}{ll}
\toprule
\textbf{Feature Extractor} & \textbf{Perturbation Strengths \( \lambda \)} \\
\midrule
CLIP ViT-L/14     & 0.05, 0.10 \\
DINOv2 ViT-L/14   & 0.05, 0.10 \\
DINOv3 ViT-S/16   & 0.05, 0.10 \\
DINOv3 ViT-H/16   & 0.05, 0.10 \\
ConvNeXT          & 0.05, 0.10 \\
BEiT ViT-L/16     & 0.05, 0.10 \\
\bottomrule
\end{tabular}
\end{table}

\subsection{Hyperparameter Configurations}
\label{appendix:hyperparams}

The following hyperparameter values were used consistently across all experiments:

\begin{itemize}
    \item Aggregation method: minimum \( p \)-value.
    \item Number of bins for pairwise \(\chi^2\) tests: \( B_{\chi^2} = 15 \).
    \item Number of bins for ECDF estimation: \( B_{\text{ECDF}} = 400 \).
    \item Significance level for KS test of uniformity: \( \alpha_{\text{KS}} = 0.05 \).
    \item Maximum Cramér’s \(V\) threshold for accepting independence in \(\chi^2\) tests: \( V_{\chi^2} = 0.07 \).
\end{itemize}

These values were chosen after exploring a range of configurations during development. Specifically, we evaluated \( B_{\text{ECDF}} \in [200, 1000] \), \( B_{\chi^2} \in [10, 50] \). We also tested thresholds up to \( V_{\chi^2} = 0.1 \) for Cramér’s \(V\) independence criterion and up to \( \alpha_{\text{KS}} = 0.1 \) for the KS uniformity test.

The final configuration was selected based on its empirical stability and the ability to maintain uniformity of aggregated \( p \)-values across different datasets and experimental setups.

\subsection{Pseudo-code}
\label{appendix:pseudocode}
\begin{algorithm}[htbp]
\caption{Null Hypothesis Modeling}
\begin{algorithmic}[1]
\STATE \textbf{Input:} Reference dataset $\mathcal{D}_{\text{real}}$
\STATE \textbf{Output:} ECDFs $\{\widehat{F}_{j,k}\}$, Subset $\mathcal{I} \subseteq \mathcal{S}$

\STATE \textit{// Statistic extraction and ECDF modeling (parallelizable)}
\FORALL{detector $(j,k)$} 
    \FORALL{$x_i \in \mathcal{D}_{\text{real}}$} 
        \STATE Compute $s_{j,k}(x_i)$ 
    \ENDFOR
    \STATE Estimate ECDF $\widehat{F}_{j,k}$ from $\{s_{j,k}(x_i)\}$
\ENDFOR

\STATE \textit{// Independence subset selection (pairwise tests are parallelizable)}
\STATE Initialize independence graph $G = (\mathcal{V}, \mathcal{E})$
\FORALL{statistic pairs $(s_i, s_j)$} 
    \STATE Compute Cramér’s $V(s_i, s_j)$ from $\chi^2$ statistic on $p$-values
    \IF{$V(s_i, s_j) \leq V_{\chi^2}$}
        \STATE Add edge $(s_i, s_j)$ to graph $G$
    \ENDIF
\ENDFOR

\STATE Find all cliques in $G$
\STATE Keep cliques passing KS-test with threshold $\alpha_{\text{KS}}$
\STATE Select valid clique maximizing coverage of preferred statistics, breaking ties by size
\end{algorithmic}
\end{algorithm}

\begin{algorithm}[htbp]
\caption{Inference}
\begin{algorithmic}[1]
\STATE \textbf{Input:} Image $x$, ECDFs $\{\widehat{F}_{j,k}\}$, subset $\mathcal{I}$, level $\alpha$
\STATE \textbf{Output:} Decision: \texttt{REAL} or \texttt{FAKE}

\STATE \textit{// Statistic computation and mapping (parallelizable)}
\FORALL{statistic $(j,k) \in \mathcal{I}$}
    \STATE Compute statistic $s_{j,k}(x)$
    \STATE Compute $p$-value $p_{j,k}(x)$ using $\widehat{F}_{j,k}$
\ENDFOR

\STATE Aggregate $\{p_{j,k}(x)\}$ into unified $p$-value (e.g., Stouffer or min-\(p\))
\STATE \textbf{if} unified $p$-value $< \alpha$ \textbf{then return} \texttt{FAKE}
\STATE \textbf{else return} \texttt{REAL}
\end{algorithmic}
\end{algorithm}

\section{Statistical Validity}
\label{appendix:proofs}

This section presents lemmas that justify the statistical validity of our hypothesis testing and $p$-value aggregation procedures. The results establish that the constructed $p$-values are valid under the null hypothesis, assuming independence and i.i.d. real samples.

\subsection*{Lemma 1: Validity of Empirical $p$-Values}

Let \( x \sim \mathbb{P}_{\text{real}} \) and let \( s(x) \) be a scalar statistic. Let \( \widehat{F}_N \) denote the ECDF over \( N \) i.i.d. samples \( x_1, \dots, x_N \sim \mathbb{P}_{\text{real}} \). Define the two-sided empirical $p$-value as
\[
p(x) = 2 \cdot \min\left( \widehat{F}_N(s(x)),\, 1 - \widehat{F}_N(s(x)) \right).
\]
Then under the null hypothesis,
\[
p \sim \mathcal{U}[0,1] \quad \text{as } N \to \infty.
\]

\textit{Note.} This follows directly from the definition of empirical $p$-values. We treat this as a definitional result.

\subsection*{Lemma 2: Stouffer Aggregation Validity}

Let \( p_1, \dots, p_K \sim \mathcal{U}[0,1] \) be independent. Define
\begin{equation}
Z = \frac{1}{\sqrt{K}} \sum_{i=1}^K \Phi^{-1}(p_i)
\end{equation}
\begin{equation}
P_{\text{Stouffer}} = \Phi(Z)
\end{equation}
where \( \Phi \) is the standard normal cumulative distribution function. Then under the null,
\[
P_{\text{Stouffer}} \sim \mathcal{U}[0,1].
\]

\paragraph{Proof.} Since each \( p_i \sim \mathcal{U}[0,1] \), the transformation \( z_i = \Phi^{-1}(p_i) \) yields \( z_i \sim \mathcal{N}(0,1) \). Independence of the \( p_i \)'s implies independence of the \( z_i \)'s. Therefore,
\[
Z = \frac{1}{\sqrt{K}} \sum_{i=1}^K z_i \sim \mathcal{N}(0,1),
\]
and thus \( P_{\text{Stouffer}} = \Phi(Z) \sim \mathcal{U}[0,1] \).
\hfill\(\square\)

\subsection*{Lemma 3: Minimum $p$-Value Aggregation Validity}

Let \( p_1, \dots, p_K \sim \mathcal{U}[0,1] \) independently, and define
\[
P_{\min} = \min_i p_i, \quad F_{P_{\min}}(t) = 1 - (1 - t)^K.
\]
Then under the null,
\[
F_{P_{\min}}(P_{\min}) \sim \mathcal{U}[0,1].
\]

\paragraph{Proof.} For any \( t \in [0,1] \),
\[
\mathbb{P}(P_{\min} \leq t) = 1 - \mathbb{P}(p_1 > t, \dots, p_K > t) = 1 - (1 - t)^K.
\]
Define the transformation
\[
U = F_{P_{\min}}(P_{\min}) = 1 - (1 - P_{\min})^K.
\]
Then for any \( u \in [0,1] \),
\begin{align*}
\mathbb{P}(U \leq u) &= \mathbb{P}\left(P_{\min} \leq 1 - (1 - u)^{1/K}\right) \\
&= F_{P_{\min}}(1 - (1 - u)^{1/K}) = u,
\end{align*}
so \( U \sim \mathcal{U}[0,1] \).
\hfill\(\square\)

\noindent These lemmas ensure that our $p$-value construction and aggregation yield statistically valid outputs under the null hypothesis, assuming independence and i.i.d. sampling from the real distribution.

\section{List of Generative Models}
\label{appendix:generators}

Our evaluation includes synthetic images produced by a diverse range of generative models, drawn from several benchmark datasets: CNNSpot, Universal Fake Detect, Stable-Diffusion-Faces, Synthbuster and GenImage. These datasets collectively span multiple generations of image synthesis techniques and architectural families.

The following models were used in our experiments: ProGAN~\cite{karras2017progressive}
, StyleGAN (WhichFaceIsReal)
, StyleGAN2~\cite{choi2020stargan}
, BigGAN~\cite{brock2018large}
, GauGAN~\cite{park2019semantic}
, CycleGAN~\cite{zhu2017unpaired}
, StarGAN~\cite{choi2018stargan}
, Cascaded Refinement Networks (CRN)~\cite{chen2017photographic}
, Implicit Maximum Likelihood Estimation (IMLE)~\cite{li2019diverse}
, Spatially-Adaptive Normalization (SAN)~\cite{dai2019second}
, DeepFake~\cite{rossler2019learning}
, Stable Diffusion (v1.4, v1.5, v2, XL)~\cite{rombach2022high, podell2023sdxl}
, ADM~\cite{li2024adm}
, VQDM~\cite{egiazarian2024accurate}
, WuKong
, Glide~\cite{nichol2021glide}
, Midjourney v5~\cite{midjourney2023v5}
, and DALL-E 3~\cite{dalle3_openai}.

As sources of real images, we relied on benchmark datasets including LSUN~\cite{yu2015lsun}, MSCOCO~\cite{lin2014microsoft}, ImageNet~\cite{deng2009imagenet}, and LAION~\cite{schuhmann2021laion}. Both our reference and test sets were constructed from these datasets, while ensuring that they consist of disjoint samples. In addition, we incorporated a large number of additional images from MSCOCO that are not part of the predefined benchmark subsets, thereby broadening the diversity and coverage of our real image pool.

\section{Full Experimental Results}
In this section, we present detailed experimental results.

\subsection{Runtime and Memory Usage Analysis}
\label{appendix:runtime_memory}

\paragraph{Runtime Under Increasing Statistics and Parallelism}
We evaluated the run-time and memory efficiency of our method across the entire phases of the framework using a representative subset of the MSCOCO dataset and SDXL fake images, 2K images total. All experiments were conducted on a multi-core CPU system and a single NVIDIA A100 80GB GPU, using parallelization with multiple workers.

To assess the scalability of our statistic extraction, we measured its runtime under different GPU worker configurations as the number of selected statistics increases.
\begin{figure}[htbp]
    \centering
    \includegraphics[width=0.6\textwidth]{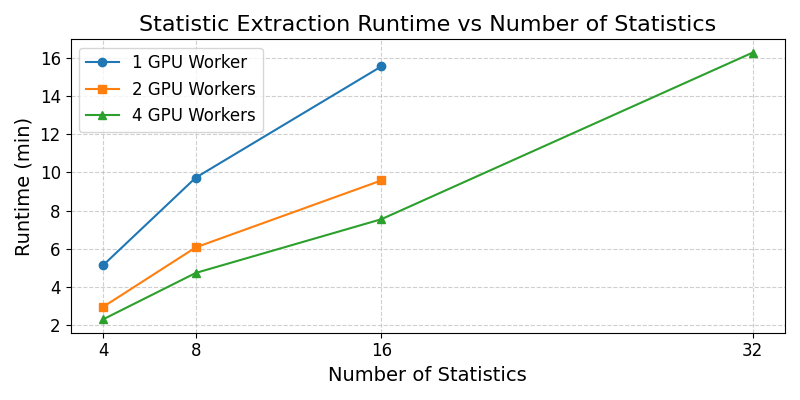}
    \caption{Statistic extraction runtime versus number of statistics under different levels of parallelism. Our method scales efficiently across multiple GPU workers.}
    \label{fig:statistic_extraction_runtime}
\end{figure}
Figure~\ref{fig:statistic_extraction_runtime} confirm that while runtime naturally increases with the number of statistics, our method benefits greatly from parallelism. For instance, with 16 statistics, runtime drops from over 15 minutes with 1 worker to under 8 minutes with 4 workers. At inference time, the per-sample runtime remains low: processing 2000 samples with 4 statistics on a single worker takes only 5.5 minutes, which corresponds to \textbf{0.165 seconds per sample}. This is substantially faster than ManifoldBias and AEROBLADE, which require 2.4 seconds and 5.1 seconds per sample, respectively.

To ensure fair comparison, we limited our evaluation to a single GPU, as baseline methods do not support multi-GPU or multi-process execution. Our implementation, however, is designed for scalability: it distributes computation across multiple workers, reducing CPU-GPU context-switch overhead. Each worker can operate on a separate GPU, compute statistics independently, and return compact ECDFs to the CPU, enabling efficient deployment in both single and multi-GPU environments.

\paragraph{Empirical Complexity Analysis of Independence-Selection Phase}
Next, we assess the cost of independence testing and clique selection.
\begin{figure}[htbp]
    \centering
     \includegraphics[width=0.6\textwidth]{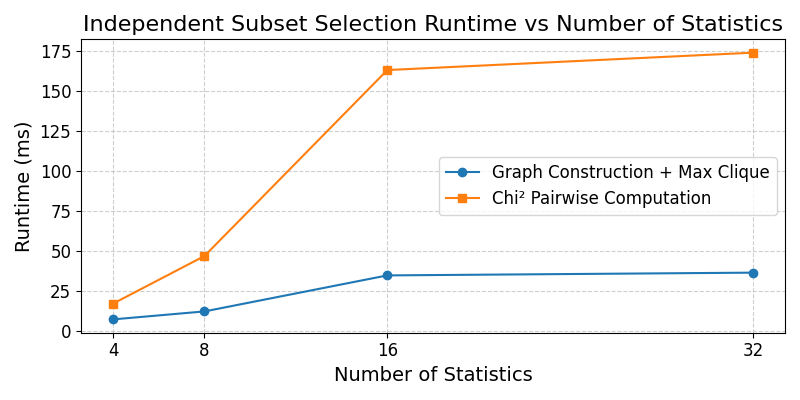}
    \caption{Runtime of statistical independence analysis and max-clique selection as the number of statistics increases. Both components scale efficiently.}
    \label{fig:clique_runtime_analysis}
\end{figure}
Figure~\ref{fig:clique_runtime_analysis} demonstrates that both the pairwise \(\chi^2\) matrix computation and the graph-based clique selection remain tractable, even with 32 statistics, requiring under 200ms end-to-end. This supports the practical deployment of our method.

Building on this analysis, we now examine the behaviour of the independence-selection stage under a wider range of conditions, including less favourable regimes that may produce denser independence graphs.

\paragraph{Empirical Complexity for Denser Independence Graphs}

To further assess computational efficiency, we conducted an experiment analysing the behaviour of maximal-clique enumeration within the independence-selection stage. Although maximal-clique search is NP-hard in theory, its practical cost is strongly influenced by the structure of the independence graph. In our settings, two characteristics ensure that this stage remains lightweight:  
\begin{itemize}
    \item independence graphs contain a relatively small number of statistics (up to 32 in all evaluated configurations).
    \item Cramer’s~V pruning produces sparse connectivity patterns before clique extraction.
\end{itemize}
As a result, clique enumeration consistently requires only tens of milliseconds in real executions, as shown in Figure~\ref{fig:clique_runtime_analysis}.  

To examine scalability under less favourable conditions, we simulated Phase~1, Stage~2 using synthetic statistics (200k values each, normalised to $[0,1]$ with small perturbations) while varying the number of candidate statistics. For each configuration, pairwise Cramer’s~V values were computed, the corresponding independence graph was constructed, and Bron-Kerbosch enumeration was performed. The recorded runtimes are summarised in Table~\ref{tab:clique_scaling}. Even under artificially dense configurations, runtime grows smoothly with graph size and remains within practical limits.

\begin{table}[hbt]
\centering
\small
\caption{Scaling behaviour of the independence-selection stage under synthetic dense conditions.}
\label{tab:clique_scaling}
\begin{tabular}{c|c|c}
\hline
\textbf{No. of Stats} & \textbf{Cramer’s V (ms)} & \textbf{Graph + clique enumeration (ms)} \\
\hline
8   & 160.3   & 0.52   \\
16  & 692.3   & 1.55   \\
32  & 2855.2  & 2.73   \\
64  & 11514.9 & 8.26   \\
128 & 46504.9 & 44.21  \\
\hline
\end{tabular}
\end{table}

These findings show that, despite worst-case theoretical complexity, the independence graphs encountered in practice keep maximal-clique extraction computationally negligible relative to feature computation and statistic evaluation. Consequently, this stage does not present a scalability bottleneck within the RealStats framework.

\paragraph{Memory Consumption and Scalability}
We further evaluate peak GPU memory usage under different configurations and compare it to the baselines.
\begin{figure}[H]
    \centering
    \includegraphics[width=0.6\textwidth]{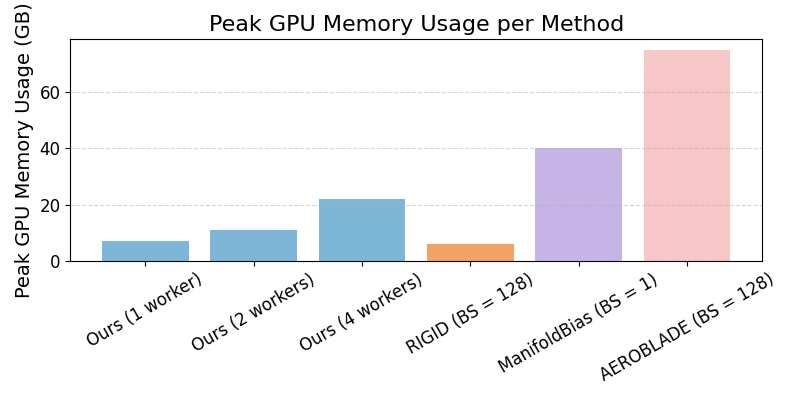}
    \caption{Peak GPU memory usage comparison with specified batch size. Our method, achieves high throughput with significantly reduced memory requirements.}
    \label{fig:peak_gpu_memory}
\end{figure}
Figure~\ref{fig:peak_gpu_memory}  shows that our method requires just 7-22GB of GPU memory with 1-4 workers at batch size 128. In contrast, AEROBLADE uses 76GB under the same setup, while ManifoldBias, with 8 internal perturbations, consumes 40GB at batch size 1 and cannot scale to larger batches. Despite using multiple statistics, our method remains memory-efficient and scalable. ECDF storage is also minimal, just 0.25MB for 32 statistics, making the approach practical even on constrained hardware.

Overall, these results demonstrate that our method is not only statistically principled, but also computationally efficient and offering fast inference, strong scalability with GPU workers, and memory usage that is well within practical bounds.

\subsection{Robustness to Image Corruption}
\label{appendix:robustness_corruptions}

We evaluate robustness to two common real-world image degradations: JPEG compression and Gaussian blur. These corruptions are applied at test time without modifying the original reference distribution, computed ECDFs, or selected statistics \( \mathcal{I} \). All experiments are conducted on the full dataset setup described in Section~C, using the entire test set of both real and fake samples to ensure comprehensive evaluation.

Importantly, metrics are aggregated over all samples, rather than averaged across splits, so that results directly reflect the overall distributional behavior. 

We evaluate robustness to two common degradations:
\begin{itemize}
    \item \textbf{JPEG Compression} (Quality Factor = 0.75) 
    \item \textbf{Standard Gaussian Blur} (3x3 Kernel)
\end{itemize}

\begin{table}[htb]
    \centering
    \small
    \caption{Performance under image corruptions for both ensemble strategies.}
    
    \label{tab:corruption_auc}
    \begin{tabular}{lcccccc}
        \toprule
        \multirow{2}{*}{\textbf{Metric}} & \multicolumn{3}{c}{\textbf{Stouffer}} & \multicolumn{3}{c}{\textbf{MinP}} \\
        \cmidrule(lr){2-4} \cmidrule(lr){5-7}
         & \textbf{Without} & \textbf{JPEG} & \textbf{Blur} & \textbf{Without} & \textbf{JPEG} & \textbf{Blur} \\
        \midrule
        AUC & 0.775 & 0.736 & 0.798 & 0.776 & 0.735 & 0.794 \\
        AP  & 0.798 & 0.750 & 0.818 & 0.786 & 0.734 & 0.807 \\
        \bottomrule
    \end{tabular}
\end{table}

Across both ensemble strategies, JPEG compression leads to a moderate reduction in performance (roughly 5-6\% drop in AUC and AP), but does not induce a significant distributional shift in the $p$-values of real samples, as the null hypothesis remains valid. Gaussian blur introduces minor changes and even improves performance slightly, suggesting that the method is robust to mild smoothing effects regardless of the ensemble variant.

\begin{figure}[H]
\centering

\begin{subfigure}[b]{0.31\linewidth}
    \includegraphics[width=\linewidth]{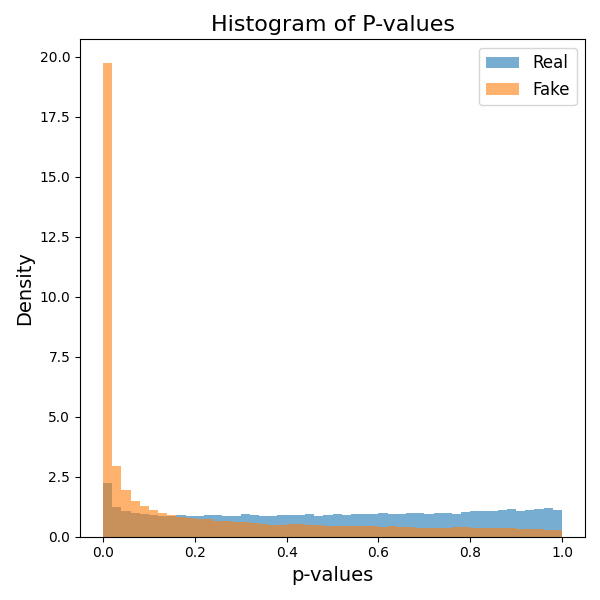}
    \caption{Stouffer: Before corruption}
\end{subfigure}
\hfill
\begin{subfigure}[b]{0.31\linewidth}
    \includegraphics[width=\linewidth]{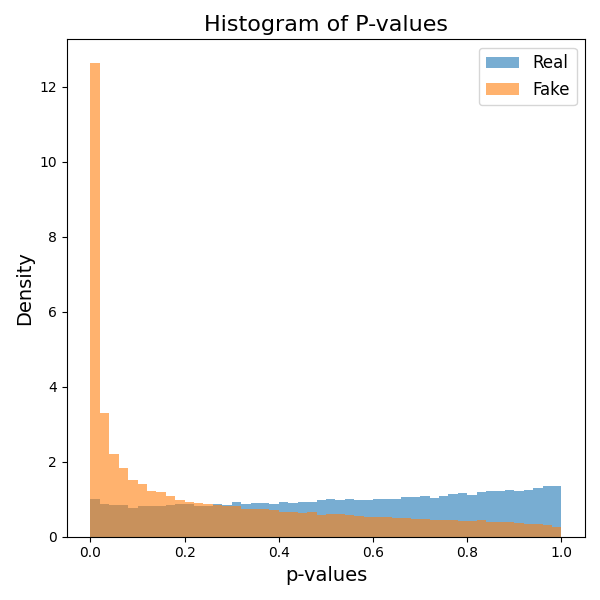}
    \caption{Stouffer: After JPEG}
\end{subfigure}
\hfill
\begin{subfigure}[b]{0.31\linewidth}
    \includegraphics[width=\linewidth]{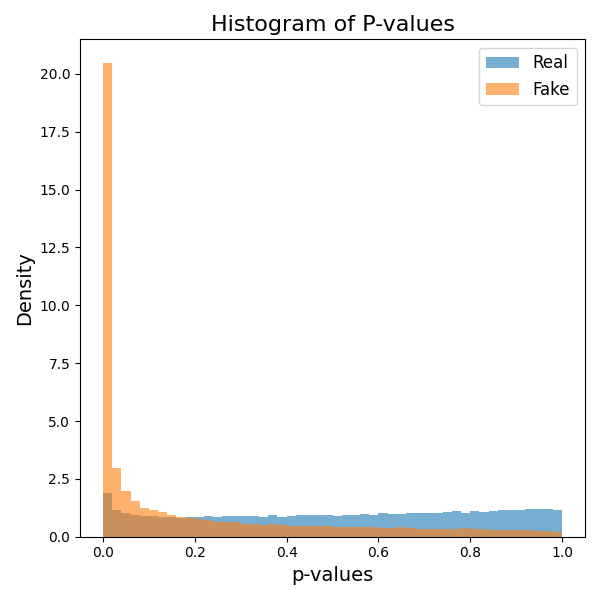}
    \caption{Stouffer: After blur}
\end{subfigure}

\begin{subfigure}[b]{0.31\linewidth}
    \includegraphics[width=\linewidth]{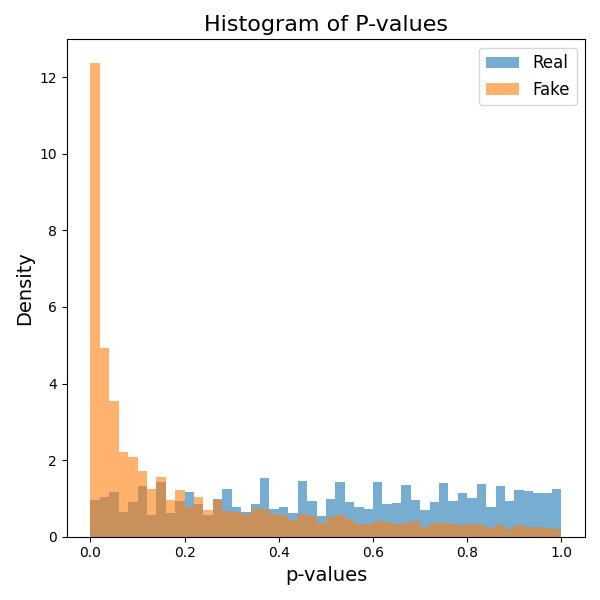}
    \caption{Min-$p$: Before corruption}
\end{subfigure}
\hfill
\begin{subfigure}[b]{0.31\linewidth}
    \includegraphics[width=\linewidth]{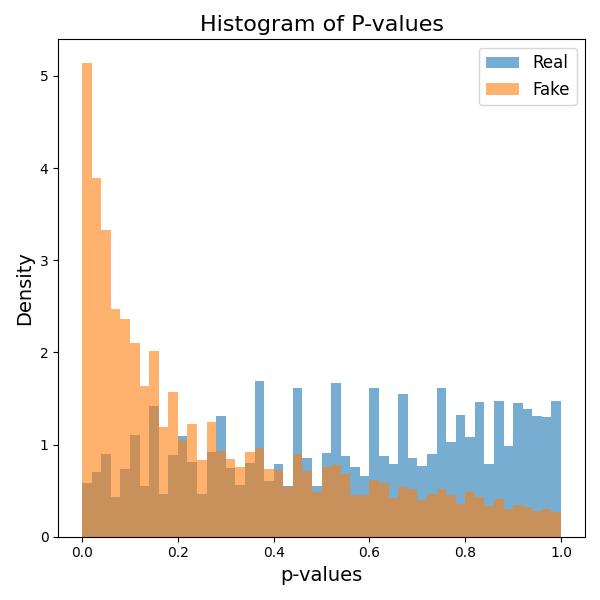}
    \caption{Min-$p$: After JPEG}
\end{subfigure}
\hfill
\begin{subfigure}[b]{0.31\linewidth}
    \includegraphics[width=\linewidth]{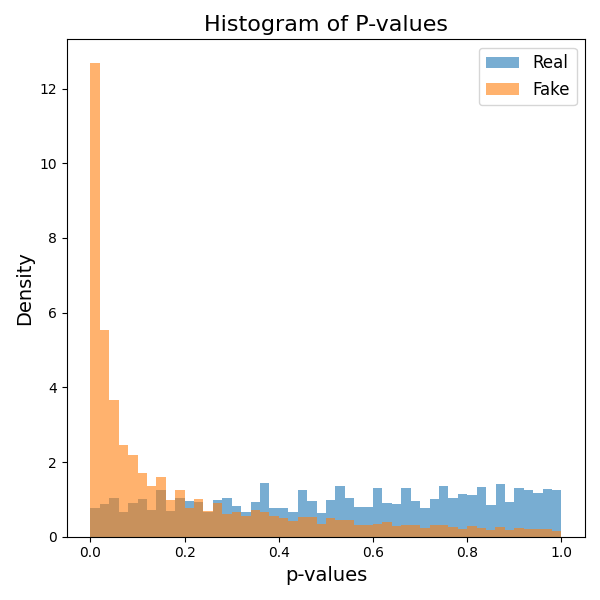}
    \caption{Min-$p$: After blur}
\end{subfigure}

\caption{
$p$-value distributions for real (blue) and fake (orange) samples under the full dataset setup. 
\textbf{Top row (Stouffer)}: Before corruption (left), after JPEG compression (middle), and after Gaussian blur (right). 
\textbf{Bottom row (Min-$p$)}: Before corruption (left), after JPEG compression (middle), and after Gaussian blur (right). 
Across both ensemble variants, JPEG introduces only a mild shift reducing separability, while Gaussian blur leaves distributions well aligned with the reference, consistent with the observed negligible or positive performance changes.
}
\label{fig:pval_shift_corruptions}
\end{figure}

Overall, these results show that our method remains statistically sound under realistic corruptions: JPEG compression introduces only moderate degradation without violating the null hypothesis, while mild Gaussian blur leaves performance unchanged or slightly improved. The robustness trends are consistent across both Stouffer and Min-$p$ ensembles.

\subsection{Preliminary Adversarial Perturbation Analysis}

We conduct a focused preliminary experiment to examine the behaviour of RealStats under simple gradient-based adversarial perturbations. This analysis is not intended as a comprehensive adversarial robustness evaluation, but rather as an initial diagnostic aligned with the statistical structure of the framework.

We consider an FGSM-style iterative perturbation targeting a single perturbation-stability statistic (DINO-based RIGID configuration). For each selected fake image, we estimate the real-image reference distribution of the statistic and apply gradient steps designed to move the statistic toward the empirical mean of the real distribution. The perturbation is applied directly in pixel space, while all ECDFs, independence selection, and aggregation procedures remain fixed.

Across multiple samples and perturbation strengths, we did not observe cases in which the unified aggregated $p$-value increased toward the real range. Instead, perturbed samples consistently shifted toward lower $p$-values, resulting in stronger confidence of being classified as fake. This behaviour was consistent across all tested configurations.

We hypothesise that this effect arises from two aspects of the framework: (i) the use of two-sided ECDF-based $p$-values calibrated on real-image distributions, and (ii) aggregation across multiple statistics that are empirically validated to satisfy independence under the null. A full adversarial robustness analysis is left for future work.

\subsection{Interpretability Qualitative Comparison}
\label{appendix:interpretability_qualitative}
\begin{figure}[htb]
    \centering
    \includegraphics[width=\textwidth]{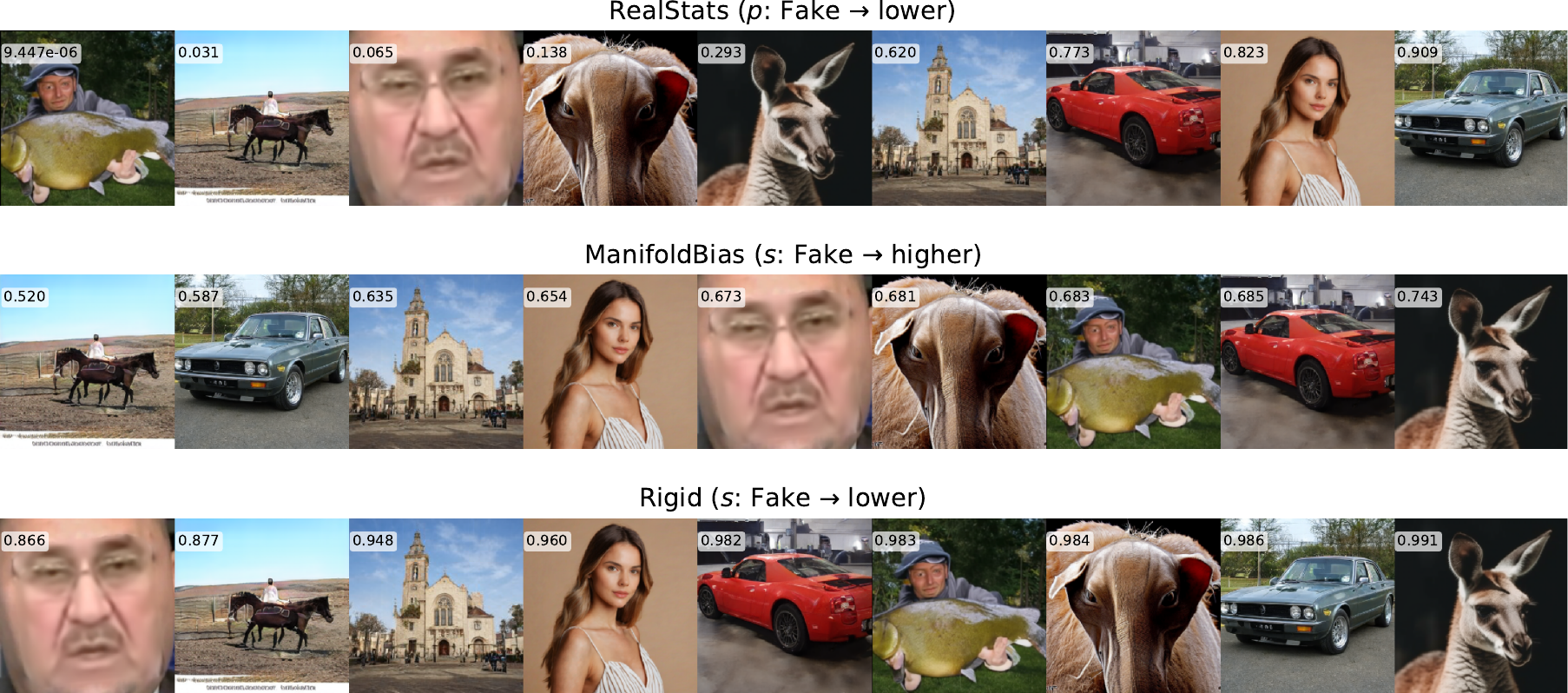}
    \caption{
    \textbf{Qualitative comparison of interpretability across methods.} 
        Each row shows scores assigned by a different method to the same set of fake images. RealStats produces $p$-values that vary consistently with deviation from the reference distribution, while the baselines often cluster visually diverse samples into similar scores.
    }
    \label{fig:method_interpretability_comparison}
\end{figure}
\begin{figure}[htb]
\centering
\includegraphics[width=1.0\textwidth]{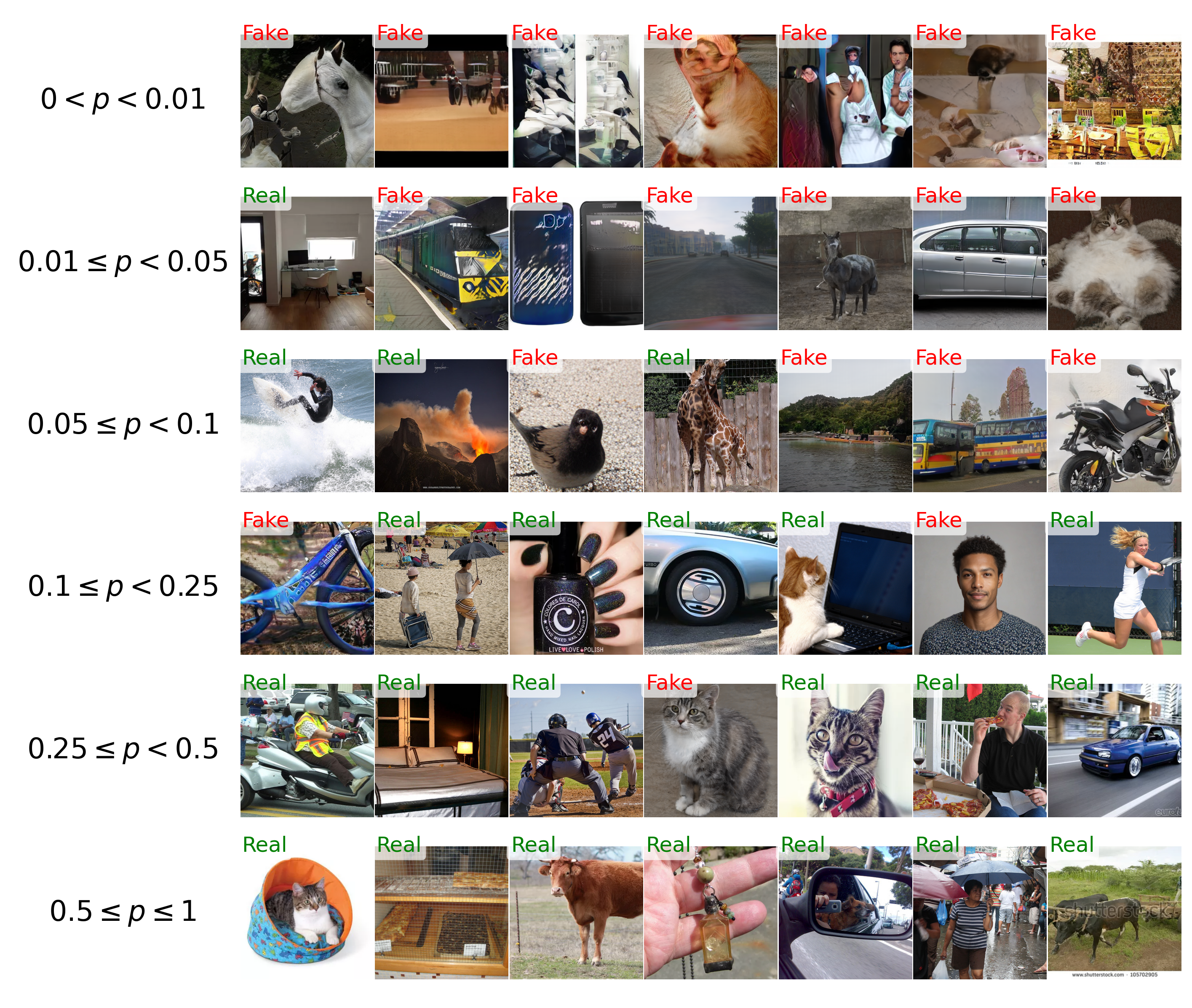}
\caption{
\textbf{Qualitative examples of $p$-value trends on test data.}
Images are grouped by $p$-value intervals (low to high). Fake images concentrate at low $p$-values, while a few overlap with real images at higher values, showing statistical similarity to the reference distribution.
}
\label{fig:pvalues_distribution}
\end{figure}
In Figure~\ref{fig:method_interpretability_comparison} we provide a qualitative comparison of interpretability across methods using a shared set of synthetic reference images. Our approach RealStats outputs calibrated $p$-values that quantify image-wise deviation from the reference distribution in an interpretable statistical framework. The resulting scores follow a progression that corresponds to the degree of visual realism, with lower $p$-values indicating a stronger appearance of being fake.

In contrast, the baseline methods, ManifoldBias and Rigid, produce scores whose distances are difficult to interpret. Images with severe visual defects may receive values close to realistic samples, and images from the same semantic domain (e.g., faces) often collapse to similar scores despite substantial corruptions. This lack of calibration limits interpretability, as the scores fail to reflect meaningful distances between samples.

Overall, these results highlight the benefit of using reference-calibrated statistics: while RealStats may not always perfectly separate fake from real, it provides an interpretable scale for reasoning about deviations from the reference distribution.

We further illustrate this interpretability by showing $p$-value distributions on our test data (Figure~\ref{fig:pvalues_distribution}). Each row corresponds to a different $p$-value interval, with images annotated by ground-truth labels. Fake images tend to cluster at lower $p$-values, though some appear at higher values, indicating statistical similarity to real images. This visualization highlights how $p$-values provide a calibrated and interpretable measure of sample quality.  

\clearpage
\subsection{Full Table of AUC and AP Values}
\label{appendix:full_auc_ap}
Table~\ref{tab:full_auc_ap_bolded} reports the complete per-generator AUC and AP values across all evaluated methods. 
These detailed results complement the averaged scores reported in the paper (Table 1), providing a comprehensive view of method performance for each generator. 
As explained, individual baselines show strong variability across generators while our method maintains competitive and more consistent results overall. 

\begin{table}[!h]
\centering
\caption{AUC and AP scores for each generator across methods.}

\resizebox{\textwidth}{!}{%
\begin{tabular}{lcccccccc}
\toprule
\multirow{2}{*}{Generator} & \multicolumn{4}{c}{AUC} & \multicolumn{4}{c}{AP} \\
 & Ours (Min-$p$) & RIGID & AEROBLADE & ManifoldBias & Ours (Min-$p$) & RIGID & AEROBLADE & ManifoldBias \\
\midrule
ADM & 0.690 & 0.575 & \textbf{0.873} & 0.611 & 0.691 & 0.590 & \textbf{0.845} & 0.564 \\
BigGAN & 0.800 & 0.794 & 0.733 & \textbf{0.909} & 0.773 & 0.798 & 0.747 & \textbf{0.918} \\
CRN & \textbf{0.967} & 0.941 & 0.845 & 0.941 & \textbf{0.930} & 0.924 & 0.836 & 0.915 \\
CycleGAN & 0.785 & 0.787 & 0.706 & \textbf{0.973} & 0.790 & 0.779 & 0.721 & \textbf{0.977} \\
DALLE & 0.552 & 0.588 & \textbf{0.710} & 0.709 & 0.552 & 0.562 & \textbf{0.689} & 0.692 \\
deepfake & 0.926 & \textbf{0.974} & 0.779 & 0.600 & 0.886 & \textbf{0.970} & 0.764 & 0.569 \\
GauGAN & 0.823 & 0.932 & 0.705 & \textbf{0.983} & 0.790 & 0.930 & 0.683 & \textbf{0.983} \\
GLIDE\_100\_27 & 0.920 & \textbf{0.978} & 0.962 & 0.911 & 0.889 & \textbf{0.978} & 0.939 & 0.923 \\
GLIDE\_50\_27 & 0.929 & \textbf{0.982} & 0.964 & 0.920 & 0.905 & \textbf{0.982} & \textbf{0.981} & 0.920 \\
IMLE & \textbf{0.963} & 0.953 & 0.807 & 0.950 & \textbf{0.942} & 0.940 & 0.829 & 0.933 \\
Midjourney & 0.661 & 0.563 & \textbf{0.692} & 0.584 & \textbf{0.661} & 0.531 & 0.705 & 0.564 \\
ProGAN & 0.851 & 0.926 & 0.878 & \textbf{0.957} & 0.845 & 0.923 & 0.902 & \textbf{0.964} \\
SAN & 0.572 & 0.655 & 0.611 & \textbf{0.778} & 0.592 & 0.668 & 0.625 & \textbf{0.780} \\
SDv1.4 & 0.745 & 0.415 & 0.421 & \textbf{0.740} & \textbf{0.720} & 0.423 & 0.444 & 0.646 \\
SDv1.5 & 0.764 & 0.394 & 0.403 & \textbf{0.763} & \textbf{0.750} & 0.414 & 0.391 & 0.674 \\
SDv2 & 0.631 & \textbf{0.820} & 0.450 & 0.587 & 0.623 & \textbf{0.820} & 0.463 & 0.635 \\
SDXL & 0.644 & \textbf{0.853} & 0.612 & 0.534 & 0.605 & \textbf{0.853} & 0.597 & 0.565 \\
StarGAN & 0.876 & \textbf{0.887} & 0.577 & 0.268 & 0.847 & 0.845 & 0.561 & \textbf{0.367} \\
StyleGAN2 & 0.576 & \textbf{0.762} & 0.606 & 0.645 & 0.545 & \textbf{0.747} & 0.584 & 0.685 \\
VQDM & 0.774 & 0.791 & 0.701 & \textbf{0.852} & 0.774 & \textbf{0.809} & 0.718 & 0.848 \\
whichfaceisreal & 0.809 & \textbf{0.946} & 0.823 & 0.744 & 0.769 & \textbf{0.925} & 0.846 & 0.722 \\
Wukong & 0.751 & 0.403 & 0.482 & \textbf{0.751} & \textbf{0.751} & 0.420 & 0.468 & 0.720 \\
\midrule
Average & 0.773 & 0.769 & 0.697 & 0.760 & 0.756 & 0.765 & 0.697 & 0.753 \\
Std & 0.126 & 0.194 & 0.161 & 0.178 & 0.119 & 0.189 & 0.163 & 0.169 \\
\bottomrule
\end{tabular}%
}
\label{tab:full_auc_ap_bolded}
\end{table}

\clearpage
\subsection{Adaptability Scores with ManifoldBias}
\label{appendix:adaptability_manifoldbias}
The results in Table~\ref{tab:adaptability_full} demonstrate the impact of incorporating adaptability into our ensemble methods. 
By allowing the aggregation strategy to adjust dynamically to different generators, our approach achieves consistently stronger performance, yielding notable improvements in both AUC and AP compared to our base model.
This highlights the importance of adaptability as a key factor for generalization across diverse generative models.

\begin{table}[!h]
\centering
\small
\caption{Per-generator AUC and AP scores of ensemble methods when integrated with ManifoldBias, with improvements (in \%). Bold indicates positive improvement.}
\begin{tabular}{lcccc}
\toprule
\multirow{2}{*}{Generator} & \multicolumn{2}{c}{AUC} & \multicolumn{2}{c}{AP} \\
 & Min-$p$ & Stouffer & Min-$p$ & Stouffer \\
\midrule
ADM & 0.644 (-6.67\%) & 0.610 & 0.641 (-7.24\%) & 0.594 \\
BigGAN & \textbf{0.869 (+8.62\%)} & 0.865 & \textbf{0.869 (+12.42\%)} & 0.862 \\
CRN & \textbf{0.974 (+0.72\%)} & 0.990 & \textbf{0.956 (+2.80\%)} & 0.986 \\
CycleGAN & \textbf{0.905 (+15.29\%)} & 0.913 & \textbf{0.914 (+15.70\%)} & 0.917 \\
DALLE & \textbf{0.584 (+5.80\%)} & 0.558 & \textbf{0.603 (+9.24\%)} & 0.570 \\
DeepFake & \textbf{0.934 (+0.86\%)} & 0.917 & \textbf{0.903 (+1.92\%)} & 0.886 \\
GauGAN & \textbf{0.949 (+15.31\%)} & 0.936 & \textbf{0.957 (+21.14\%)} & 0.939 \\
Glide\_100\_27 & \textbf{0.941 (+2.28\%)} & 0.951 & \textbf{0.940 (+5.74\%)} & 0.957 \\
Glide\_50\_27 & \textbf{0.948 (+2.05\%)} & 0.962 & \textbf{0.945 (+4.42\%)} & 0.966 \\
IMLE & \textbf{0.973 (+1.04\%)} & 0.993 & \textbf{0.953 (+1.17\%)} & 0.991 \\
Midjourney & 0.518 (-21.63\%) & 0.518 & 0.526 (-20.42\%) & 0.523 \\
ProGAN & \textbf{0.942 (+10.69\%)} & 0.936 & \textbf{0.951 (+12.54\%)} & 0.946 \\
SAN & \textbf{0.670 (+17.13\%)} & 0.668 & \textbf{0.688 (+16.22\%)} & 0.663 \\
SDXL & \textbf{0.646 (+0.31\%)} & 0.645 & \textbf{0.608 (+0.50\%)} & 0.622 \\
SDv2 & \textbf{0.722 (+14.42\%)} & 0.708 & \textbf{0.704 (+13.00\%)} & 0.682 \\
SDv1.4 & 0.689 (-7.52\%) & 0.679 & 0.679 (-5.69\%) & 0.660 \\
SDv1.5 & 0.707 (-7.46\%) & 0.710 & 0.705 (-6.00\%) & 0.706 \\
StarGAN & \textbf{0.887 (+1.26\%)} & 0.864 & \textbf{0.871 (+2.83\%)} & 0.821 \\
StyleGAN2 & \textbf{0.609 (+5.73\%)} & 0.615 & \textbf{0.604 (+10.83\%)} & 0.647 \\
VQDM & \textbf{0.793 (+2.45\%)} & 0.781 & \textbf{0.784 (+1.29\%)} & 0.766 \\
WhichFaceIsReal & \textbf{0.820 (+1.36\%)} & 0.765 & \textbf{0.802 (+4.29\%)} & 0.743 \\
WuKong & 0.707 (-5.86\%) & 0.709 & 0.702 (-6.52\%) & 0.706 \\
\midrule
Average & 0.792 & 0.786 & 0.787 & 0.780 \\
Std & 0.143 & 0.151 & 0.141 & 0.151 \\
\bottomrule
\end{tabular}
\label{tab:adaptability_full}
\end{table}

\section{Datasets and Code}
\label{appendix:reproducibility}
We release both the \textbf{data splits} used in our experiments and the full \textbf{codebase}. 
The package includes our implementations of \textbf{RIGID statistics} alongside the other statistical measures evaluated in this work, as well as scripts for inference and the complete pipeline. 
We provide the exact \textbf{random seeds} used in all benchmarks and evaluations: \{0, 8, 12, 18, 22, 28, 30, 32, 36, 38\}. 

All resources (code and datasets) are available in our GitHub repository: \url{https://github.com/shaham-lab/RealStats}.


\thispagestyle{empty}
\end{document}